# LAG-XAI: A Lie-Inspired Affine Geometric Framework for Interpretable Paraphrasing in Transformer Latent Spaces

Olexander Mazurets[1], Olexander Barmak[*,1], Leonid Bedratyuk[1], Iurii Krak[2,3]

[1] Department of Computer Science, Khmelnytskyi National University, Khmelnytskyi, 29016, Ukraine
[2] Department of Theoretical Cybernetics, Taras Shevchenko National University of Kyiv, Kyiv, 03187, Ukraine
[3] Laboratory of Communicative Information Technologies, V.M. Glushkov Institute of Cybernetics of NAS of Ukraine, Kyiv, 03187, Ukraine

Emails: mazuretso@khmnu.edu.ua, barmako@khmnu.edu.ua, leonidbedratyuk@khmnu.edu.ua, iurii.krak@knu.ua

*Corresponding author: Olexander Barmak (barmako@khmnu.edu.ua)

ORCID:
Olexander Mazurets: https://orcid.org/0000-0002-8900-0650
Olexander Barmak: https://orcid.org/0000-0003-0739-9678
Leonid Bedratyuk: https://orcid.org/ 0000-0002-6076-5772
Iurii Krak: https://orcid.org/ 0000-0002-8043-0785

**ABSTRACT**
Modern Transformer-based language models achieve strong performance in natural language processing tasks, yet their latent semantic spaces remain largely uninterpretable black boxes. This paper introduces LAG-XAI (Lie Affine Geometry for Explainable AI), a novel geometric framework that models paraphrasing not as discrete word substitutions, but as a structured affine transformation within the embedding space. By conceptualizing paraphrasing as a continuous geometric flow on a semantic manifold, we propose a computationally efficient mean-field approximation, inspired by local Lie group actions. This allows us to decompose paraphrase transitions into geometrically interpretable components: rotation, deformation, and translation. Experiments on the noisy PIT-2015 Twitter corpus, encoded with Sentence-BERT, reveal a "linear transparency" phenomenon. The proposed affine operator achieves an AUC of 0.7713. By normalizing against random chance (AUC 0.5), the model captures approximately 80% of the non-linear baseline's effective classification capacity (AUC 0.8405), offering explicit parametric interpretability in exchange for a marginal drop in absolute accuracy. The model identifies fundamental geometric invariants, including a stable matrix reconfiguration angle (~27.84°) and near-zero deformation, indicating local isometry. Cross-domain generalization is confirmed via direct cross-corpus validation on an independent TURL dataset. Furthermore, the practical utility of LAG-XAI is demonstrated in LLM hallucination detection: using a "cheap geometric check," the model automatically detected 95.3% of factual distortions on the HaluEval dataset by registering deviations beyond the permissible semantic corridor. This approach provides a mathematically grounded, resource-efficient path toward the mechanistic interpretability of Transformers.

## 1. INTRODUCTION

Paraphrase variability – the ability to express identical meaning through diverse syntactic and lexical structures – is a highly complex object of study in natural language processing (NLP) [1]. Modern NLP systems based on the Transformer architecture yield high-quality sentence embeddings, enabling the effective comparison of texts in high-dimensional latent spaces [2]. However, despite progress in computing semantic similarity, such models remain largely opaque black boxes. The underlying nature of the geometric trajectories connecting semantically equivalent sentences (paraphrases) within these spaces is largely hidden from researchers.

There is a fundamental discrepancy between the linguistic complexity of paraphrasing and the existing methods for its mathematical modeling. On the one hand, detailed linguistic taxonomies (such as Bhagat and Hovy's classification) identify over 25 types of quasi-paraphrases, including lexical substitutions, syntactic reformulations, and active-to-passive voice alternations [3]. On the other hand, in most practical NLP applications, this multifaceted process is reduced to a single static scalar – typically cosine similarity or Euclidean distance [4]. This approach ignores the geometry of the transition trajectory in the embedding space, treating it as an uninformed spatial displacement. Consequently, it limits the explainability regarding exactly how the model transforms one sentence into another and whether the original meaning is preserved.

Furthermore, evaluating generative models currently incurs substantial computational costs. To verify paraphrasing quality or factual accuracy (e.g., hallucination detection), the "LLM-as-a-judge" approach is frequently employed, which requires resource-intensive queries to larger models (such as GPT-4). This constrains real-time verification in high-load systems. There is an emerging need for efficient AI methods capable of performing reliable semantic verification through fast "cheap geometric checks" directly within the latent space, without requiring the generation of new tokens.

The interpretability challenge is particularly pronounced when processing specific social media data. For instance, in the PIT-2015 (Twitter) corpus [5], paraphrases are noisy, concise, and often involve a superposition of multiple linguistic transformations. This necessitates the development of an explainable artificial intelligence (XAI) operator capable of modeling the logic of language models through transparent linearization mechanisms [6], bridging complex theoretical geometric frameworks with practical, computationally tractable solutions.

This paper approaches the problem through the lens of geometric deep learning [7] and mechanistic interpretability. We hypothesize that the semantic manifold of paraphrases is locally isometric and that the paraphrasing process can be mathematically described as a local affine transformation $T(x) = Ax + t$ acting on the embedding space. Drawing inspiration from the framework of Lie groups and algebras [8], our approach extends beyond merely establishing text similarity; it allows interpreting paraphrase transitions as local geometric transformations, thereby enabling the extraction of empirically dominant deformation directions that function as approximate generators of the underlying transformation group. Decomposing this XAI operator into rotational, deformational, and translational components enables the mathematical separation of structural changes (syntax) from pragmatic shifts (meaning) [9].

The main contributions of this work are as follows:

- **Demonstrating the phenomenon of "linear transparency"**: A structurally consistent affine model is developed, showing that a significant portion (~80%, normalized above random chance) of the effective variability in Transformer model decisions (using SBERT as an example) is linear in nature. We formalize this trade-off between accuracy and interpretability, where a marginal absolute reduction in AUC (~7%) is compensated by the capacity for mechanistic analysis of the transformations.
- **Identification of geometric language invariants**: The study identifies geometric constants of the paraphrasing process. It demonstrates that this process is locally isometric (occurring with near-zero deformation of information volume) and is characterized by a stable structural reconfiguration angle of the basis.
- **Introduction of XAI descriptors and the concept of efficient AI validation**: A system of linguistically interpretable metrics is developed, including the observed effect of "semantic chirality." These XAI descriptors can function as fast out-of-distribution (OOD) anomaly detectors, offering a reliable alternative to resource-intensive methods for evaluating generation quality and detecting hallucinations.

This research provides a mathematical foundation for shifting from opaque vector comparisons toward controlled text generation, where exact geometric modifications of embeddings "on the fly" along the identified isometric trajectories can replace computationally expensive data augmentation.

To evaluate the generalizability of the observed geometric properties, the method is additionally validated in a cross-domain setting on the independent Twitter URL Corpus (TURL) [16], illustrating the model's capacity to generalize paraphrasing rules without further fine-tuning.

The remainder of the paper is organized as follows: Section 2 provides a review of related work; Section 3 details the materials, mathematical foundations, and experimental methodology; Sections 4 and 5 present the results and discussion; Section 6 summarizes the conclusions and outlines directions for future research.

## 2. RELATED WORKS

The problem of paraphrase modeling has evolved from detailed linguistic taxonomies [3] to complex Transformer-based architectures [11]. Recent research in the geometric analysis of language models has focused on the local geometry of their latent spaces. Notably, recent studies in mechanistic interpretability indicate that Transformer models encode abstract concepts through linear subspaces, making affine approximation a well-founded analytical tool [12].

Despite the strong performance of Sentence-BERT [2] in semantic similarity tasks [4], recent literature highlights the issue of embedding "anisotropy" – a phenomenon where sentence vectors are confined to narrow cones, thereby skewing cosine similarity [13]. This creates a need to transition from static metrics to dynamic transformation operators, particularly when processing noisy social media data [5].

It is essential to distinguish the proposed approach from other latent space analysis methods. Traditional dimensionality reduction techniques (such as PCA or t-SNE) primarily focus on visualizing the static distribution of data points, often ignoring the vector fields that describe transitions between them. Conversely, linguistic probing methods can detect the presence of specific syntactic information within a vector but fail to explain the mechanism of its transformation (i.e., *how* the change occurs). The present study bridges this gap by modeling paraphrasing as continuous motion along a manifold, allowing not only the assertion of similarity but also the quantitative description of the semantic change "trajectory."

A further distinction must be drawn between the proposed framework and existing geometric probing methods [20]. While classical probing aims to find static linear subspaces that encode specific linguistic properties (e.g., syntax trees), our research is directed toward modeling the dynamics of transformations between states. Unlike probing methods, which merely identify the presence of features within a vector, the affine formulation via Lie groups enables the direct observation of the mechanism by which one meaning "flows" into another. This shifts interpretability from a passive state (feature detection) to an active one (analysis of operations on features), providing tools for decomposing semantic shifts into geometrically interpretable transformation components (rotation, deformation, and translation) alongside their associated XAI descriptors (reorientation angle, deformation index, and shift magnitude).

This geometric perspective aligns with recent trends in neural computing and applications, which have extensively explored both architectural modifications to Transformers and the structure of their latent spaces. For instance, Stragapede et al. (2024) demonstrated the efficacy of adapting the Transformer architecture (TypeFormer) for specific biometric tasks [17]. Our study complements this direction by offering a geometric interpretation of how attention mechanisms structure semantic space. Furthermore, the application of the "geometric corridor" concept for hallucination detection aligns with emerging approaches in anomaly detection, where deviations from a normal distribution in the latent space indicate out-of-distribution (OOD) data [18].

Recent advances in equivariant deep learning [8] and the use of Lie groups to model continuous symmetries in data [14] pave the way for the development of explainable AI (XAI) models [6]. However, as noted in foundational reviews on geometric deep learning [7], the application of Lie algebras to the high-dimensional spaces of NLP models remains underexplored. Most existing methods are oriented towards computer vision, whereas the "physics" of semantic motion in text embeddings has thus far lacked a precise mathematical description [9]. This study integrates Lie group theory with modern approaches to stabilizing matrix computations [10, 15] to address this gap.

A summary of the analyzed approaches and the positioning of our method is presented in Table 1.

**Table 1. Comparative analysis of approaches and identified gaps**

| Approach | Object of Analysis | Key References | Limitations (Gaps) |
|---|---|---|---|
| **Linguistic** | Syntactic types | [1, 3] | Ignores the geometric structure of vector spaces. Employs a descriptive rather than a computational approach. |
| **Vector (State-of-the-Art)** | Distance (Cosine/L2) | [2, 13] | The anisotropy problem; low interpretability ("black box"); static comparison without trajectory analysis. |
| **Mechanistic Interpretability** | Linear subspaces | [12, 15] | Focuses on model weights rather than value transformation operations. Frequently limited to basic probing tasks. |
| **Geometric Probing** | Static features in subspaces | [20] | Does not explain the dynamics of changes; focuses on *where* information is located rather than *how* it transforms. |
| **Lie-XAI (Our approach)** | Affine operator and its geometric decomposition | [8, 14, 17] | Requires high computational stability for 768D+ spaces. Explains the transformation mechanism through the parametric decomposition of the operator. |

Consequently, the objective of this study is to determine the geometric invariants and statistical boundaries of the transformations that describe the transition between paraphrase sentence vectors. The primary focus is on validating the hypothesis of the locally isometric nature of paraphrasing and verifying the phenomenon of "linear transparency" in complex Transformer models to ensure a high level of explainability (XAI) for semantic deformation processes.

To achieve this objective, the following formal problem statement is formulated. Consider a natural language sentence space $\mathcal{X}$ and a fixed non-linear encoder model $f: \mathcal{X} \to R^n$ (e.g., Sentence-BERT), which maps each sentence $s \in \mathcal{X}$ to a dense vector embedding $x = f(s)$ in an $n$-dimensional space.

Given a paraphrase corpus $\mathcal{P} = \{(s_i, s_i', y_i)\}_{i=1}^N$, where $s_i, s_i' \in \mathcal{X}$ represent a pair of sentences, and $y_i \in \{0,1\}$ is a binary label indicating whether $s_i'$ is a semantically equivalent paraphrase of $s_i$. For each pair, the corresponding embeddings $x_i = f(s_i)$ and $x_i' = f(s_i')$ are computed.

**Problem 1: Estimation of a structurally consistent transformation operator.**

Using a subset of true paraphrases ($I = \{i \mid y_i = 1\}$), the objective is to find the optimal principal affine operator $T(x) = Ax + t$, where $A \in R^{n \times n}$ is the feature basis linear reconfiguration matrix, and $t \in R^n$ is the contextual shift vector. The task requires not only minimizing the mean squared approximation error functional:

$$E(A, t) = \sum_{i \in I} \|x_i' - (Ax_i + t)\|_2^2, \qquad (1)$$

but also discovering the law of invariance of transformation.

Based on the spatial homogeneity hypothesis of the semantic manifold, paraphrasing is modeled as an integral motion along an invariant vector field. In this formulation, the operator $A$ serves as a mean-field approximation for the set of local tangent maps. This enables the isolation of the dominant geometric mode of semantic transformation, which remains structurally stable across the entire data distribution in the latent space.

**Problem 2: Geometric decomposition, XAI profiling, and invariant identification.**

For the derived principal operator $(A, t)$, a structural analysis must be performed using polar decomposition and Lie algebras. The task aims to mathematically separate pure basis rotation (corresponding to syntactic restructuring) from symmetric deformation (corresponding to changes in information volume). Based on this, key XAI descriptors must be computed: the semantic deformation index ($Def$), the generalized transformation angle ($\theta$), and the presence or absence of a "semantic chirality" effect (via the determinant sign). Establishing the statistical boundaries of these parameters outlines a "geometric corridor" within which a text transformation preserves its semantic identity.

Thus, the absence of a unified mathematical operator that explains paraphrasing dynamics via Lie groups underscores the need to develop an affine model that balances mechanistic interpretability with high computational efficiency.

## 3. MATERIALS AND METHODS

The methodology of this study is structured into a sequential pipeline, from raw text to an interpretable geometric decision. Figure 1 provides a high-level overview of this entire process, which encompasses data vectorization, geometric preprocessing, the core affine operator estimation, and the final XAI-driven analysis for paraphrase validation and anomaly detection. Each stage of this pipeline is detailed in the subsequent subsections.

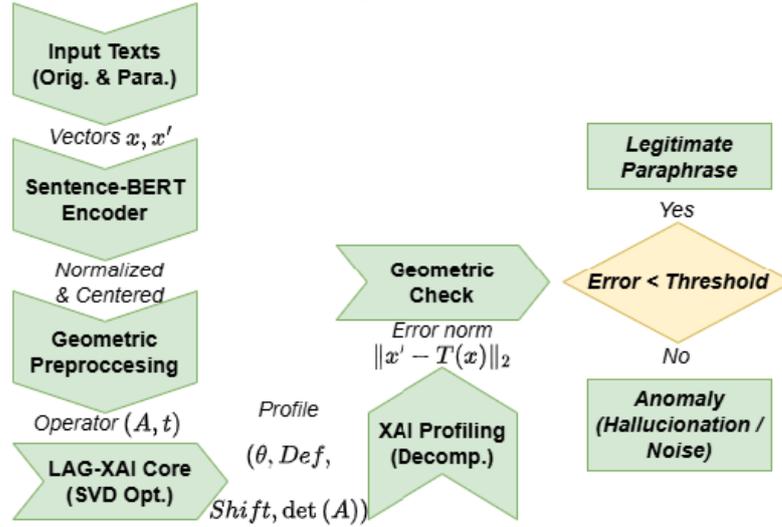

**Fig. 1**. The complete methodological pipeline of the LAG-XAI framework. The process begins with encoding input texts into dense vectors using Sentence-BERT. These vectors then undergo geometric preprocessing (L2-normalization and centering). The core LAG-XAI model solves a regularized optimization problem to find the global affine operator $(A, t)$. This operator is then decomposed to generate an interpretable XAI profile. Finally, a "geometric check" based on the approximation error determines whether the text pair is a legitimate paraphrase or a semantic anomaly, such as an LLM hallucination.

### 3.1. Theoretical Background: Semantic Manifold and the Affine Lie Group

In this work, we propose LAG-XAI (Lie Affine Geometry for Explainable AI), a framework that models semantic transformations within Transformer latent spaces via the affine actions of Lie groups. Unlike traditional approaches, LAG-XAI conceptualizes paraphrasing as a continuous geometric flow on a semantic manifold.

**Definition (LAG-XAI)**: LAG-XAI is a geometric framework that models paraphrasing as a global affine transformation $T(x) = Ax + t$, which is interpreted as a mean-field approximation of local Lie group actions on a semantic manifold. While explicit analytical computation of Lie algebras via logarithmic mapping is computationally prohibitive for high-dimensional semantic spaces ($n \geq 768$), our approach employs established algebraic techniques (PCA and Procrustes regularization) to construct a robust empirical approximation of these geometric generators.

The proposed approach is fundamentally grounded in the manifold hypothesis, which posits that high-dimensional dense embeddings generated by deep language models (e.g., SBERT) do not populate the Euclidean space $R^n$ chaotically. Instead, they form a smooth, low-dimensional semantic manifold $\mathcal{M} \subset R^n$ [7]. Within this geometric paradigm, paraphrasing – the transition from a sentence $x$ to its equivalent $x'$ – is viewed not as a random discrete jump, but as the outcome of a continuous dynamic process, representing a semantic flow along the surface of this manifold.

Since the explicit governing law linking the embedding of an original sentence $x$ to its paraphrase $x'$ is not defined analytically and may be highly non-linear, its direct analysis is computationally challenging. Therefore, we employ the principle of local linearization: the paraphrase correspondence in the embedding space is modeled as an implicit transformation $\Phi: \mathcal{M} \to \mathcal{M}$ that aligns pairs $(x, x')$ within the corpus. For thematically proximal sentences (such as paraphrases), the mapping $\Phi$ near a point $x$ can be approximated by its differential $d\Phi_x: T_x\mathcal{M} \to T_{x'}\mathcal{M}$, which acts as a linear operator on the tangent spaces. As conceptually illustrated in Fig. 2, at the coordinate level in $R^n$, this yields an affine approximation of the transition between paraphrases, expressed as:
$$x' \approx T(x) = Ax + t,$$

where the matrix $A$ corresponds to the linear part of the local action (first-order term), and the vector $t$ represents the free term (translation/shift).

It is crucial to clarify the transition from a local to a global description. In differential geometry, the differential $d\Phi_x$ generally depends on the point $x$ and can vary along the manifold $\mathcal{M}$. In our formulation, we introduce a spatial homogeneity hypothesis regarding semantic transformations within the SBERT space: we assume that for a broad class of paraphrase pairs in each corpus, there is a shared, dominant character of local deformations, and the corresponding linearizations $d\Phi_x$ are proximal to one another, with only minor contextual fluctuations. Under this hypothesis, the sought global operator $A$ is not an arbitrary statistical average but is interpreted as a stable (consensus) approximation of the typical local first-order transition $x \mapsto x'$ in the embedding space – that is, the matrix that optimally reconciles the set of local tangent maps in an average sense.

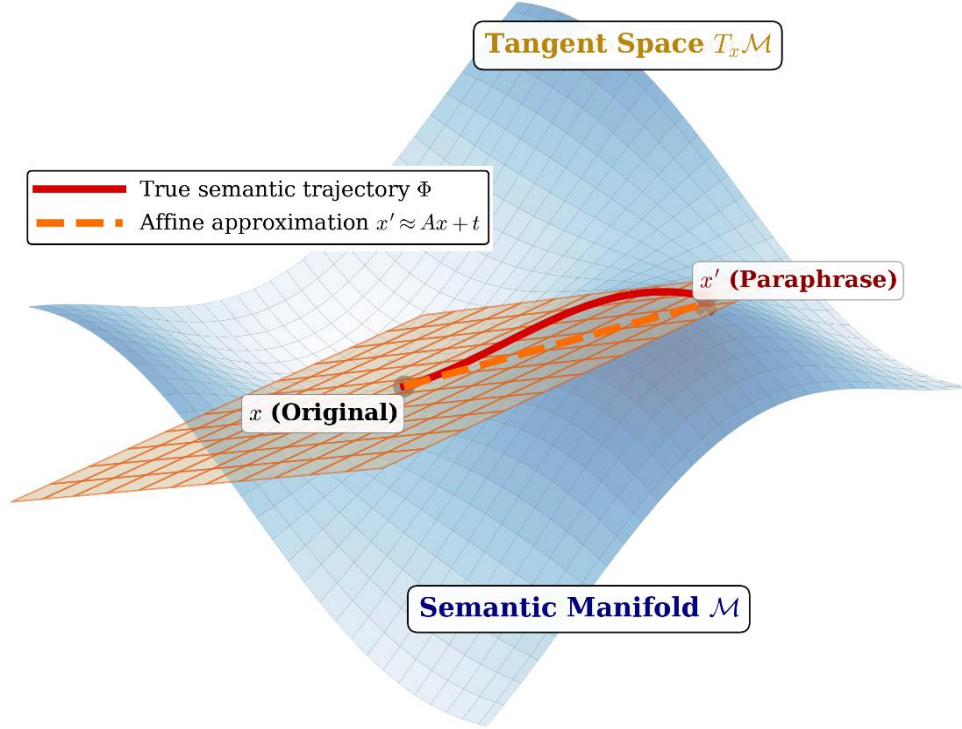

**Fig. 2.** Conceptual visualization of the LAG-XAI framework on the semantic manifold $\mathcal{M}$. The highly non-linear transformation between the original sentence $x$ and its true paraphrase $x'$ (represented by the red trajectory $\Phi$) is locally approximated within the tangent space $T_{x\mathcal{M}}$. The dashed orange vector demonstrates the linear affine mapping $x' \approx Ax + t$, which projects the semantic transition onto the linear subspace, enabling geometric interpretability.

Formally, the transition from a family of local $d\Phi_x$ to a single global operator $A$ can be interpreted as a mean-field approximation, where $A$ captures the typical action of local differentials over the set of observed paraphrases. In a practical setting, this "average" is computed with respect to the empirical distribution of points $x$ in the corpus:
$$A \approx E_{x \sim \mathbb{D}}[d\Phi_x],$$
and the precision of this approximation is characterized by the quantity:
$$\Sigma_{loc} := E_{x \sim \mathbb{D}}[\|d\Phi_x - A\|_F^2],$$
which reflects the variance of local linearizations around the global consensus. Empirically, the validity of a global description is supported by the fact that the estimated operator $A$ approximates a near-isometric transformation (see Fig. A.2b), aligning with the hypothesis of bounded local distortions in the given embedding space.

The affine group $\text{Aff}(n)$ consists of transformations of the form $x \mapsto Ax + t$, where $A \in \text{GL}(n, R)$ and $t \in R^n$, and can be represented as the semidirect product of its linear part and translations:
$$\text{Aff}(n) \cong \text{GL}(n, R) \ltimes R^n.$$

Accordingly, the transition to the paraphrase vector is described by the equation:
$$x' = Ax + t, \qquad (2)$$
where the elements of the transformation play interpretable roles:

- The matrix $A \in \text{GL}(n, R)$ acts as a linear operator governing the reorientation and deformation of the semantic feature basis; it accounts for the structural reconfiguration of the expression (e.g., changes in word order, or transitions from active to passive voice).
- The vector $t \in R^n$ defines the translation (contextual shift) and models pragmatic or stylistic alterations that displace the representation of the expression into a different spatial region.

The local structure of a Lie group is described by its Lie algebra $\mathfrak{g}$, whose elements can be interpreted as infinitesimal "directions" of change. Since language models operate as "black boxes," defining the corresponding infinitesimal objects analytically is impossible. Therefore, we introduce an empirical analog: we analyze the set of drift vectors:

$$\Delta x = x' - x,$$

computed on paraphrase pairs within the corpus and apply Principal Component Analysis (PCA) to them. The resulting dominant directions $\{p_1, \ldots, p_r\}$ are interpreted as a basis of principal semantic shift modes (at the first order) characteristic of the given corpus and embedding model. These directions are utilized as constraints/regularizers during the estimation of matrix $A$, restricting the search space to geometrically meaningful deformations consistent with the empirically observed paraphrasing dynamics.

Furthermore, from a linguistic perspective, ideal paraphrasing entails preserving a sentence's meaning (semantic equivalence) despite potential variations in form. Geometrically, this motivates approximating a near-isometric action in the embedding space: to ensure that operator $A$ does not distort local distances in the first approximation, it should belong (or be maximally close) to the class of orthogonal transformations, satisfying the condition:

$$A^\top A \approx I,$$

where $I$ is the identity matrix. This theoretical assumption regarding the near-isometric nature of moderate paraphrase transitions necessitates orthogonal regularization (e.g., via the Procrustes problem) during the optimization of model parameters.

The advantage of the affine formulation over existing representation-level interpretability tools lies in its generative nature. While most XAI methods (such as LIME or SHAP) rely on statistical correlations, LAG-XAI is grounded in the physical analogy of motion along a manifold. This enables not merely "highlighting" important vector dimensions but mathematically separating syntactic restructuring from semantic shift—a capability unavailable to standard subspace probing methods.

In this study, *geometric invariants* refer to scalar characteristics derived from the parameters of the affine operator $T(x) = Ax + t$ that are robust to the choice of basis in the feature space. That is, they remain invariant under orthogonal coordinate changes $x \mapsto Qx$ ($Q \in O(n)$) and consequently yield a comparable "geometric profile" of the paraphrase transition across different sentence pairs. Unlike the matrices $A$, $R$, and $U$ themselves, which are specific to particular pairs, these invariants are coordinate-stable values suitable for statistical analysis and boundary establishment.

For each paraphrase pair $(x, x')$, following the estimation of $A$ and $t$, and the polar decomposition $A = RU$, we compute the following four descriptors:

1. **Rotation index ($\theta$)**: An aggregated measure of the rotational component $R$, reflecting the intensity of structural reconfiguration within the feature space.

2. **Deformation index ($Def$)**: A measure of non-isometry derived from the singular values of matrix $A$ (equivalently, from the spectrum of the symmetric part $U$ in the polar decomposition), which characterizes the compression or expansion of the semantic representation.

3. **Shift magnitude ($\|t\|_2$)**: The norm of the translational component, quantitatively describing the pragmatic or contextual drift that cannot be explained solely by linear basis restructuring.

4. **Orientation indicator ($\det(A)$)**: Specifically, the sign of $\det(A)$, which determines whether the linear part of the transformation includes an orientation-reversing component (reflection).

These specific invariants are subsequently utilized to construct the statistical boundaries of paraphrasing and to provide an explainable interpretation of the differences between paraphrases and non-paraphrases.

### 3.2. Computational Model of Structurally Consistent Affine Linearization

While rigorous differential geometry assumes a unique transformation at each point, the use of a "global consensus operator" is a justified heuristic for Transformer spaces due to semantic coherence. Because SBERT vectors are already clustered by similarity into anisotropic cones, the operator $A$ describes the principal direction of the semantic flow, where local contextual fluctuations are smoothed out by the high dimensionality of the space.

Let a training set of true paraphrases be given as matrices of original sentences $X \in R^{N \times n}$ and their equivalents $X' \in R^{N \times n}$, where $N$ is the number of pairs and $n$ is the dimensionality of the embedding space.

Since directly solving the least squares problem to determine the operator $A$ in arbitrary high-dimensional spaces ($n \gg 1$) is an ill-posed problem, leading to poor matrix conditioning and the phenomenon of "semantic space collapse" (overfitting), we propose a stable optimization model. Conceptually, it is based on the formulation and solution of a system of geometrically regularized normal equations.

The first conceptual step is the proper separation of the linear basis restructuring (matrix $A$) from the contextual shift (vector $t$). This is achieved by transitioning to mean-centered data matrices via the subtraction of the barycenters (mean vectors) $\mu_X$ and $\mu_{X'}$ of the source and target sets, respectively:

$$X_c = X - \mu_X, \quad X'_c = X' - \mu_{X'}. \tag{3}$$

Following centering, the problem reduces to finding the operator $A$ as the solution to the matrix equation $A \cdot LHS = RHS$. A key aspect of our model involves augmenting the base covariance ($X_c^\top X_c$) and cross-covariance (($X'_c)^\top X_c$) matrices with two types of stabilization constraints that preserve the local geometry of the semantic manifold:

$$LHS = X_c^\top X_c + \lambda_{ortho} I + \lambda_{equiv} J^\top J, \tag{4}$$

$$RHS = (X'_c)^\top X_c + \lambda_{ortho} R_{prior}^\top. \qquad (5)$$

The physical interpretation of the introduced regularizers is as follows:
- **Orthogonal stabilization (Procrustes Prior)**: The term weighted by the hyperparameter $\lambda_{ortho}$ penalizes the model for deviations from pure isometry, forcefully regularizing the operator $A$ toward the orthogonal subgroup $O(n)$. The prior matrix $R_{prior} \in R^{n \times n}$ is computed as the solution to the orthogonal Procrustes problem: $\arg\min_R \|X_c R^\top - X'_c\|_F$ subject to $R^\top R = I$. This constraint prevents the uncontrolled loss or generation of information volume during the transformation.
- **Structural stabilization (PCA Generators)**: The term weighted by the hyperparameter $\lambda_{equiv}$ is responsible for aligning matrix $A$ with the principal directions of paraphrasing. The matrix $J \in R^{r \times n}$ contains the first $r$ principal components (PCA) extracted from the semantic deformation matrix $\Delta X = X'_c - X_c$. In the context of our Lie-inspired framework, these principal directions serve as computationally tractable, empirical approximations of the Lie algebra generators for the local tangent space. The use of $J^T J$ as a regularizer restricts the search space, compelling the operator to act predominantly along these empirically discovered directions.

A distinct challenge in estimating the operator $A$ is due to the extreme multicollinearity in the input embeddings. Due to the anisotropy effect ("cone-shaped" distribution) inherent to the SBERT latent space, the covariance matrices often exhibit exceedingly high condition numbers ($\kappa(LHS) > 10^{15}$), rendering standard matrix inversion numerically unstable. The implemented geometric regularization system (Eq. 4–5) acts as a spectral stabilizer: the addition of the terms $\lambda_{ortho} I$ and $\lambda_{equiv} J^\top J$ effectively "lifts" the matrix's eigenvalues above the noise threshold, thereby reducing the condition number to values suitable for reliable gradient computation and pseudoinversion.

To avoid numerical instabilities during the inversion of the $LHS$ matrix caused by embedding multicollinearity, the solution is obtained via Truncated Singular Value Decomposition (Truncated SVD) with a noise truncation threshold $\tau$. Singular values smaller than $\tau$ are zeroed, acting as a low-frequency semantic noise filter.

After computing the pseudoinverse matrix $LHS^+$, the linear operator and the final translation vector are reconstructed using the following formulas:

$$A = RHS \cdot LHS^+, \qquad (6)$$
$$t = \mu_{x'} - A\mu_x. \qquad (7)$$

The computational sequence of the described concept, optimized for matrix operations, is formalized in Algorithm 1.

**Algorithm 1. Computation of the Structurally Consistent Affine Operator**

**Input**:
- $X, X' \in R^{N \times n}$ – embedding matrices (source sentences and their paraphrases);
- $\lambda_{ortho}, \lambda_{equiv}$ – weights for orthogonal and structural regularization;
- $r$ – number of principal components for empirical deformation directions;
- $\tau$ – truncation threshold for singular values (e.g., $10^{-3}$).

**Output**:
- $A \in R^{n \times n}$ – linear transformation matrix (basis restructuring);
- $t \in R^n$ – contextual translation (shift) vector.

1. **Centering**:
   Compute the barycenters $\mu_X, \mu_{X'}$ and the centered matrices $X_c, X'_c$ according to (3).
2. **Priors Estimation**:
   $R_{prior} \leftarrow \text{OrthogonalProcrustes}(X_c, X'_c)$
   $J \leftarrow \text{PCA}(X'_c - X_c, \text{n\_components} = r)$
3. **System Formulation**:
   Construct the normal equation matrices $LHS$ and $RHS$ according to (4) and (5).
4. **Regularized SVD Solver**:
   $U, \Sigma, V^\top \leftarrow \text{SVD}(LHS)$
   $\Sigma^+ \leftarrow \text{PseudoInverse}(\Sigma, \text{threshold} = \tau)$
   $LHS^+ \leftarrow V \Sigma^+ U^\top$
5. **Reconstruction**:
   Compute the affine parameters $A$ and $t$ according to (6) and (7).
6. **Return**: $(A, t)$

It is important to note that step 4 of the algorithm is utilized to mitigate multicollinearity. Zeroing singular values below the threshold $\tau = 10^{-3}$ guarantees that the effective condition number of the solution will not exceed the threshold $1/\tau = 1000$. This ensures the numerical robustness of the method even when processing very short or highly correlated sentences.

A significant advantage of the proposed method is its efficiency compared to iterative neural network optimization techniques. The computational core of the algorithm relies on the Singular Value Decomposition (SVD), which is required for solving the orthogonal Procrustes problem and the PCA regularization. For a training set of $N$ sentence pairs and embedding dimensionality $n$, the complexity of formulating the covariance matrices is $\mathcal{O}(N \cdot n^2)$, while the spectral decomposition of the $n \times n$ matrix is $\mathcal{O}(n^3)$. Thus, the overall time complexity of the algorithm is estimated as $\mathcal{O}(Nn^2 + n^3)$.

Since the dimensionality $n$ (e.g., 768 or 1024) is a fixed constant, the algorithm scales linearly with respect to the corpus size $N$. This renders the method suitable for Real-Time XAI applications, unlike approaches based on gradient descent or direct LLM queries.

### 3.3. Geometric Decomposition and XAI Profiling

The principal affine operator $(A, t)$ obtained in the previous step captures the complete physics of the semantic transformation; however, in its raw form, the $n \times n$ matrix remains opaque for linguistic analysis. To ensure the model's interpretability within the framework of explainable artificial intelligence (XAI), we employ the mathematical formalism of matrix decomposition.

Conceptually, any continuous linear transformation $A \in \mathrm{GL}(n, R)$ can be uniquely decomposed into structural components via polar decomposition:
$$A = R \cdot S, \tag{8}$$
where: the orthogonal matrix $R$ ($R^\top R = I$) accounts for pure multidimensional rotation (the reorientation of the semantic basis, which linguistically models a change in the form of expression – such as syntactic restructuring or word order alteration – while fully preserving semantic intensity, i.e., vector length); the symmetric positive-definite matrix $S$ ($S = S^\top > 0$) acts as a deformation tensor. It contains information regarding the stretching or compression of the semantic space along principal axes, which linguistically corresponds to a change in the information density of the sentence (e.g., expansion or simplification of content).

The linguistic interpretation of each component in this decomposition is visually summarized in Fig. 3.

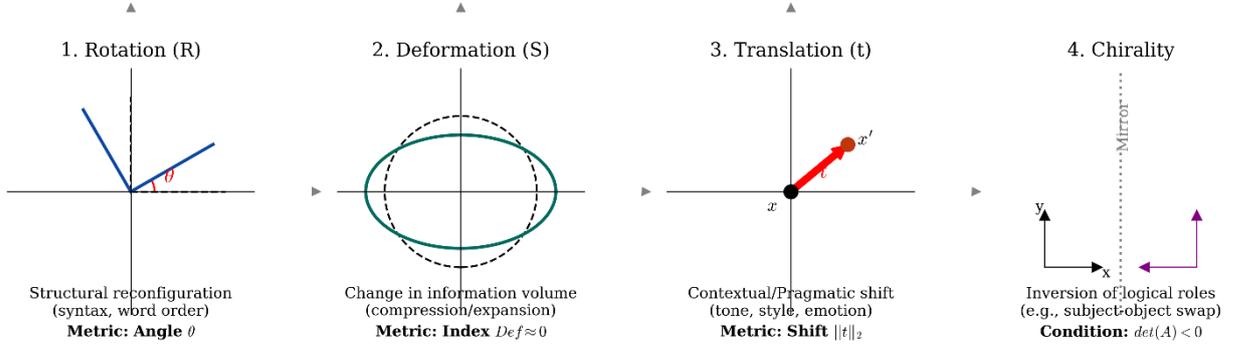

**Fig. 3.** Conceptual diagram of the LAG-XAI geometric decomposition. The affine operator $A$ is broken down into four linguistically interpretable components. **(1) Rotation (R)** models syntactic changes like word order. **(2) Deformation (S)** represents changes in information volume. **(3) Translation (t)** captures pragmatic shifts in context or tone. **(4) Chirality (det(A) < 0)** indicates a reflection, corresponding to a deep logical inversion of semantic roles.

Based on this decomposition, alongside the singular value decomposition of matrix $A$ (whose singular values coincide with the eigenvalues of matrix $S$), we construct a generalized geometric XAI profile of the paraphrase, comprising four key descriptors:

**1. Generalized structural reconfiguration angle (θ).**

This parameter quantitatively describes the intensity of basis reorientation in the latent space. Within our geometric paradigm, we consider θ as a mathematical proxy for the structural reconfiguration of the vector. Although the direct correspondence between basis rotation and linguistic syntax requires further validation through metrics such as Tree Edit Distance, high values of θ indicate a significant transformation of the internal representation structure, typically accompanying a change in word order or grammatical construction. The angle is computed based on the trace of the rotation matrix (the sum of its diagonal elements, $\mathrm{Tr}(R)$). For a space of arbitrary dimensionality $n$, the generalized transformation angle is defined as:
$$\theta = \arccos\left(\max\left[-1, \min\left(1, \frac{\mathrm{Tr}(R) - n + 2}{2}\right)\right]\right). \tag{9}$$

The larger the value of θ, the more strongly the paraphrase structure differs from the original (e.g., during a transition from active to passive voice).

**2. Semantic deformation index ($Def$).**

This descriptor quantifies the degree of non-isometry of the transformation. It is calculated as the mean absolute deviation of the singular value spectrum $\{\sigma_k\}_{k=1}^{n}$ of matrix A from unity:
$$Def = \frac{1}{n}\sum_{k=1}^{n}|\sigma_k - 1|. \tag{10}$$

This metric is crucial for testing the manifold hypothesis: the approximation $Def \to 0$ serves as a mathematical indicator that the paraphrase is locally isometric (preserving meaning without adding or removing information details).

**3 Semantic chirality (Spatial orientation).**

The geometric nature of the operator is verified by computing the sign of its determinant, $\det(A)$. If $\det(A) > 0$, the transformation preserves spatial orientation. If $\det(A) < 0$, the matrix $A$ includes a reflection operation. The presence of a negative determinant indicates the phenomenon of "semantic chirality." In this context, chirality is understood as the geometric property of an object that is not superimposable on its mirror image (an orientation-reversing isometry). This points to a deep inversion of certain feature basis components, which linguistically may correspond to a shift in logical emphasis or grammatical voice while maintaining overall semantic equivalence.

**4. Contextual shift magnitude ($Shift$).**

The translation vector $t$ models semantic displacement that is not explained by the linear basis restructuring. Its magnitude is estimated using the Euclidean norm:

$$Shift = \|t\|_2. \tag{11}$$

This descriptor reflects the pragmatic shift inherent, for example, in social media texts, where a paraphrase might be accompanied by a change in the author's attitude, emotional tone, or a transition to a new thematic context.

The combination of the computed parameters ($\theta, Def, Shift, \det(A)$) forms a unified geometric XAI profile, enabling the quantitative classification of texts based on the types of their linguistic transformations.

The procedure for automatically extracting these metrics is formalized in Algorithm 2.

**Algorithm 2. Geometric Decomposition and XAI Profile Generation**

**Input**:
- $A \in R^{n \times n}$ – linear transformation matrix (from Algorithm 1);
- $t \in R^n$ – contextual shift vector (from Algorithm 1);
- $n$ – dimensionality of the embedding space (e.g., 768).

**Output**:
- $Profile = \{\theta, Def, Shift, \det(A)\}$ – set of geometric XAI descriptors.

1. **Polar Decomposition**:
   $R, S \leftarrow \text{Polar}(A)$ according to (8).
2. **Structural reconfiguration angle computation ($\theta$)**:
   Compute the angle $\theta$ in radians according to formula (9).
   $\theta \leftarrow \text{Degrees}(\theta)$ // Conversion to degrees for interpretation
3. **Information deformation estimation ($Def$)**:
   $\{\sigma_k\}_{k=1}^n \leftarrow \text{SingularValues}(A)$ // Alternatively, the eigenvalues of $S$
   Compute the index $Def$ according to (10).
4. **Semantic chirality estimation (Orientation)**:
   $\det\_A \leftarrow \det(A)$ // If $< 0$, a reflection operation is present
5. **Contextual shift estimation ($Shift$)**:
   Compute the norm $Shift$ according to (11).
6. **Return**: $Profile = \{\theta, Def, Shift, \det(A)\}$

### 3.4. Data, Vectorization, and Geometric Normalization

To empirically validate the developed mathematical model and test the linear transparency hypothesis, the Paraphrase and Semantic Similarity in Twitter (PIT-2015) corpus [5] was selected.

The selection of this specific dataset is motivated by the idiosyncratic nature of Twitter texts. Unlike academic or news corpora (e.g., MRPC or STS-B), tweets are characterized by high levels of lexical noise, slang, atypical syntax, and extreme contextual brevity. This creates ideal conditions for stress-testing semantic transformation models, as paraphrasing in such environments is rarely a simple synonym substitution; rather, it frequently demands a deep structural reconfiguration of the expression.

**Dataset Formation and Annotation**:

The corpus contains pairs of tweets collected based on shared topics. In the original dataset, annotation was performed by human experts on a similarity scale from 0 to 5. In accordance with standard methodology for binary classification and the configuration of our experimental pipeline, the labels were binarized: pairs with a consensus score of $\geq 3$ were classified as true paraphrases ($y_i = 1$), while the rest were designated as non-paraphrases ($y_i = 0$).

The present study utilizes the following subsets:
- **Train set**: 13,063 sentence pairs, utilized exclusively for estimating the principal affine operator $(A, t)$ (Algorithm 1).
- **Dev set**: 4,727 sentence pairs, serving as the test set for XAI profiling (Algorithm 2) and evaluating the classification capacity of the model.

**Additional Corpus for Cross-Domain Validation (TURL)**:

To mitigate the risk of overfitting to the specific annotation idiosyncrasies of PIT-2015, the independent Twitter URL Corpus (TURL) [16] was incorporated into the experimental design. Unlike PIT-2015, which was manually annotated by experts, TURL was automatically generated by grouping tweets that reference identical news URLs. This data collection approach ensures natural ("in-the-wild") paraphrase variability, ranging from simple headline copying to substantive news summarization by users.

This study employs a randomly sampled subset of this corpus. This dataset was entirely excluded from the training phase (the estimation of matrix A) and was utilized exclusively for qualitative cross-domain case studies to observe the behavior of the XAI profile on out-of-domain structures.

**Corpus for Evaluating Factual Distortion (Hallucination) Detection**:

In addition to the PIT-2015 and TURL corpora, a subset of the HaluEval dataset [19] is used to evaluate the model's robustness to factual distortions (hallucinations). Unlike paraphrase datasets, this corpus contains "correct answer vs. hallucination" pairs, enabling the stress-testing of the model's geometric tolerance under conditions of factual consistency violation.

**Vectorization (Sentence Embeddings)**:

To map texts into the vector space, the Sentence-BERT model (SBERT all-mpnet-base-v2) was utilized, yielding an output vector dimensionality of $n = 768$. The selection of this architecture (as opposed to modern decoder-only LLMs such as LLaMA-3 or Mistral) is driven by the need for isolated testing of geometric properties in a controlled environment, where vector representations are the target product of representation learning rather than a byproduct of token generation. This circumvents the noise introduced by varying pooling strategies (e.g., last-token vs mean pooling) inherent to generative models.

Regarding spatial dimensionality, we posit a universal scaling assumption: the identified geometric invariants (isometry, stable angle) remain valid for spaces of varying dimensions (e.g., $n = 1024$ or $n = 4096$), provided the model is sufficiently over-parameterized to form a smooth manifold, in accordance with the hypothesis outlined in [7].

**Geometric Preprocessing ($L_2$-Normalization)**:

A critically important preprocessing step, without which the application of Lie algebraic formalism would be mathematically unsound, is the mandatory $L_2$-normalization of the generated embeddings. For each vector $x \in R^{768}$, its normalized counterpart is computed as:

$$x_{norm} = \frac{x}{\|x\|_2 + \epsilon}, \qquad (12)$$

where $\epsilon = 10^{-9}$ is a constant introduced to ensure numerical stability.

This operation carries profound geometric significance: it projects all sentences onto the surface of the unit hypersphere $S^{n-1}$. By virtue of this normalization:

1. It eliminates the influence of amplitude deviations (differences in vector norms), which SBERT may generate for sentences of varying lengths.

2. For normalized vectors, the squared Euclidean distance becomes a monotonic function of cosine similarity:
$$\| x - y \|_2^2 = 2(1 - \cos(x, y)), \text{ if } \| x \|_2 = \| y \|_2 = 1,$$
rendering these two metrics strictly equivalent up to a monotonic transformation.

3. Orthogonal transformations $R \in O(n)$ emerging from the polar decomposition of the operator's linear part preserve the norm and thus describe a "pure" reorientation of the embeddings on the sphere without altering their length. This perfectly aligns with the interpretation of the rotational component as the near-isometric constituent of the paraphrase transition.

$L_2$-normalization does not render the affine model globally spherically invariant (as the translational component $t$ generally does not preserve the sphere); however, it significantly stabilizes the angular geometry of the data and makes the interpretation of the orthogonal component $R$ mathematically sound and comparable across different pairs.

### 3.5. Theoretical Generalization and Experimental Design

*3.5.1. Applied Perspectives of Geometric Linearization*

The formalization of paraphrasing via transformation operators and their approximate infinitesimal generators not only resolves the issue of local space alignment but also delineates the following strategic research directions:

- **Development of geometrically equivariant Transformer architectures**: The proposed approach enables the integration of the identified generator $J$ directly into the neural network's computational graph as a geometric regularizer. This facilitates the construction of a model that is paraphrase-invariant "by design," as it learns to preserve the semantic core along trajectories defined by the vector field $\exp(\epsilon J)$. This method circumvents the need for resource-intensive textual data augmentation (e.g., back-translation, synonym substitution) by replacing it with continuous, on-the-fly geometric modifications to embeddings.

- **Controlled paraphrase generation and stylization**: Utilizing the $(A, t)$ decomposition allows for the isolation of components responsible for style (rotation θ) and content (translation $t$). This paves the way for controlled text synthesis, enabling users to specify the intensity of a sentence's structural reconfiguration without altering its semantic content.

- **Anomaly detection and robustness to perturbations**: Establishing the geometric boundaries of paraphrasing enables the use of the approximation error norm $\|x' - T(x)\|_2$ as an indicator of a "semantic departure beyond the manifold." Within our experimental design, this proposition is validated on the HaluEval dataset, where LLM hallucinations are treated as structural anomalies that violate local isometry. This approach is critically important for detecting adversarial attacks and automatically identifying factual distortions in generative models without relying on resource-intensive external classifiers.

- **XAI verification of semantic drift**: The proposed geometric profile ($\theta, Def, Shift, \det(A)$) provides a transparent linguistic interpretation for each transformation. The employment of the independent Twitter URL Corpus (TURL) serves to confirm the stability of these geometric invariants in a cross-corpus transfer setting, demonstrating that the estimated operator captures the fundamental physics of the latent space rather than merely the specificities of the training set. This capability enables auditing of language models' decision-making logic for structural and contextual alterations.

While these outlined directions constitute the focus of our future research, their realization necessitates the rigorous experimental validation of the foundational model, which is presented in the current study.

### 3.5.2. Evaluation Metrics and Hybrid Similarity

The primary objective of the experiment is to quantify the extent to which our linear affine abstraction can explain the complex non-linear logic of the SBERT model. To this end, we developed the Hybrid Cosine Score metric.

The evaluation process for a test pair $(x, x')$ consists of the following steps:

1. **Application of the global operator** to the source vector: $x_{pred} = Ax + t$.
2. **Projection of the theoretical result** onto the hypersphere: $x_{norm} = x_{pred}/\|x_{pred}\|_2$.
3. **Computation of the dot product**: $Score = x_{norm} \cdot x'$.

It should be emphasized that in the case where the operator $A$ is the identity matrix ($I$) and the translation vector is $t = 0$, the Hybrid Cosine Score is mathematically identical to the standard cosine similarity between the original vectors $x$ and $x'$. Thus, standard cosine similarity (SBERT Baseline) serves as a natural baseline for ablation analysis, and any deviation of the hybrid score from the baseline is strictly attributable to the action of the geometric operator $A$, rather than being a mere normalization effect.

**Scope of Evaluation and Baselines**: It is critical to contextualize the evaluation strategy. The primary objective of LAG-XAI is *not* to compete with state-of-the-art non-linear classifiers (such as cross-encoders, NLI-based models, or BERTScore) in absolute paraphrase detection accuracy. Instead, LAG-XAI is proposed as an Explainable AI (XAI) probing framework designed to open the "black box" of *existing* vector spaces. Consequently, comparing our linear operator against complex, uninterpretable neural classifiers would be a category error. The only appropriate baseline is the uninterpretable standard cosine similarity of the base model itself (SBERT in this study). This comparison directly quantifies the trade-off: how much absolute accuracy is sacrificed to gain explicit mechanistic interpretability (the $\theta, Def$, and $Shift$ metrics).

Classification performance (the ability to distinguish paraphrases from non-paraphrases) is evaluated using the Area Under the Receiver Operating Characteristic Curve (ROC-AUC). The comparison of our model's $AUC_{hybrid}$ with the baseline $AUC_{baseline}$ (standard cosine similarity between the original SBERT vectors) serves as a quantitative measure of the space's "linear transparency."

To confirm the statistical robustness of the developed metrics, the evaluation is conducted across three distinct scenarios:

1. **In-domain validation (PIT-2015)**: Assessing the hybrid metric's capacity to replicate SBERT's logic on noisy social media texts.
2. **Cross-domain validation (TURL)**: Verifying the stability of $AUC_{hybrid}$ on an independent corpus to confirm that the metric reflects fundamental properties of the latent space, rather than overfitting to a specific set of topics.
3. **Anomaly detection audit (HaluEval)**: Utilizing the approximation error norm $\|x' - T(x)\|_2$ as a specialized metric to differentiate factual hallucinations. In this scenario, the $AUC$ is calculated based on the model's ability to separate legitimate paraphrasing trajectories from anomalous "hallucinatory" transitions.

### 3.5.3. Experimental Design and Scenarios

To comprehensively evaluate the model, the experimental design (the results of which are presented in Section 4) is structured into six sequential stages:

- **Stage 1. Trade-off Optimization and Structural Analysis (Grid Search)**: Conducting a grid search on the Train set within the hyperparameter space: $\lambda_{ortho} \in [100, 5000], \lambda_{equiv} \in [0.1, 1.0], r \in [3,5]$. The objective is to identify the optimal configuration that maximizes AUC while preserving isometry ($Def \rightarrow 0$). For the derived global operator, the singular value spectrum and the distribution of translation vectors are analyzed.
- **Stage 2. Classification Capacity Validation (XAI Verification)**: Constructing ROC curves on the Dev set to compare the affine model against the baseline. Investigating the distribution of approximation errors (Residual Error) for paraphrase and non-paraphrase classes to confirm that true paraphrases form a dense cluster with minimal deviations from the principal trajectory.
- **Stage 3. Spectrum of Geometric Complexity**: Computing individual transformation angles $\theta$ for all pairs in the Dev set to visualize the "geometric corridor." Identifying extreme cases (minimum and maximum angles) for linguistic XAI auditing, ranging from simple lexical substitutions to deep syntactic restructuring.
- **Stage 4. Global vs. Local Dynamics and Cross-Domain Robustness (TURL)**: This stage evaluates the generalizability of the affine operator. The assessment is conducted across four internal scenarios based on PIT-2015:

- *Scenario A (Global - Dev)*: Global model evaluated on novel (unseen) topics.
- *Scenario B (Local - Dev)*: Local (topic-specific) models evaluated on novel topics.
- *Scenarios A.1 and B.1*: Theoretical accuracy limits on known data.

To conclusively validate the universality of the discovered laws, an external cross-domain validation is performed on the independent TURL corpus. This allows comparing the operator's efficiency across texts with fundamentally different syntactic structures (news headlines versus conversational tweets).

- **Stage 5. 3D/2D Visualization of the Semantic Manifold ("Affine Tube")**: Constructing a three-dimensional model of the semantic transition, where the axes represent the structural transformation angle (θ), the deviation error from the ideal axis (Residual Error), and cosine similarity (Cosine). This visually demonstrates the physical boundaries of the space within which a paraphrase remains legitimate.
- **Stage 6. Cross-Corpus Detection of Semantic Anomalies (HaluEval)**: The final stage of the design verifies the method's applicability to AI Safety tasks. Utilizing the HaluEval dataset (QA split), we test the capacity of the "cheap geometric check" to distinguish between factual statements and LLM hallucinations by detecting their departure beyond the boundaries of the constructed "affine tube."

## 4. RESULTS

In accordance with the outlined experimental design, computations were performed on the PIT-2015 corpus using SBERT (all-mpnet-base-v2) embeddings. The results regarding the identification and validation of the geometric invariants of paraphrasing are detailed below.

### 4.1. Trade-off Optimization and Invariant Identification (Grid Search)

In the first stage, a grid search was conducted on the training set (13,063 pairs) to identify the optimal hyperparameters for the principal affine operator. The primary objective was to achieve a balance between classification performance (ROC-AUC) and geometric stability (minimization of deformation, $Def$). A subset of the optimization results is presented in Table 2.

**Table 2. Optimal trade-off search results on the validation set.**

| Orthogonality weight ($\lambda_{ortho}$) | No. of components ($r$) | Generator weight ($\lambda_{equiv}$) | AUC (Hybrid) | Rotation angle ($\theta$) | Deformation ($Def$) |
|---|---|---|---|---|---|
| 100.0 | 5 | 1.00 | 0.7686 | 27.79° | 0.01005 |
| 500.0 | 5 | 1.00 | 0.7709 | 27.84° | 0.00235 |
| 1000.0 | 5 | 1.00 | 0.7712 | 27.84° | 0.00121 |
| **5000.0** | **5** | **1.00** | **0.7713** | **27.84°** | **0.00025** |
| *SBERT Baseline* | - | - | *0.8405* | - | - |

An analysis of the data indicates that the highest classification performance is achieved by the configuration with the maximum weight applied to the orthogonal regularizer ($\lambda_{ortho} = 5000$). Furthermore, two fundamental phenomena are observed under these conditions:

- **Local isometry**: As orthogonal stabilization increases, the semantic deformation index sharply decreases toward zero ($Def = 0.00025$). This demonstrates that the semantic manifold of paraphrases is locally isometric – the transformation preserves the informational volume of the sentence while reorienting its basis.
- **Geometric constant of the operator**: Despite variations in the regularization parameters, the generalized structural change angle of the global matrix A ($\theta_A$) exhibits strong consistency, converging at 27.84° (with a standard deviation of only 0.0219°). This allows us to assert the identification of a "geometric signature" for the affine transformation specific to the given language model.

Additionally, the calculated determinant of the optimal matrix is $\det(A) \approx -0.836$. This negative value mathematically proves the presence of a reflection operation. Such a finding indicates the "semantic chirality" of the space: a paraphrase is not a simple multidimensional rotation; it frequently encompasses a profound inversion of specific linguistic axes (e.g., the transition from active to passive voice or a shift in logical emphasis).

To isolate the contribution of the geometric operator $A$, an ablation study was conducted. Upon the removal of matrix $A$ (i.e., setting $A = I, t = 0$), the model reverts to the standard cosine similarity metric ($AUC = 0.8405$). The application of the computed operator $A$ results in an $AUC$ reduction to 0.7713. This confirms that operator $A$ does not merely exploit the $L_2$-normalization effect to artificially inflate metrics; rather, it genuinely models (approximates) the complex non-linear logic of SBERT, sacrificing approximately 8% of classification capacity in exchange for the full parametric interpretability of the transformation.

To rigorously prove that these discovered geometric invariants (specifically θ and $Def$) are intrinsic properties of the embedding space rather than artifacts of a specific dataset split, an extended bootstrap resampling analysis is detailed in **Appendix A (Fig. A.6)**. The violin plots confirm that the variance of these parameters across multiple training subsets is practically negligible, ensuring strict methodological reproducibility.

## 4.2. Structural Analysis of the Affine Operator

To verify the mathematical validity of the derived linear transformation matrix $A$ and the translation vector $t$, their physical properties were analyzed. The matrix exhibited full rank (Rank = 768), indicating the absence of information collapse; no semantic dimensions of the SBERT latent space were lost.

The Frobenius norm of the matrix ($\|A\|_F = 27.7064$) and the Euclidean norm of the translation vector ($\|t\|_2 = 0.3840$) demonstrate that the principal contribution to paraphrasing is driven by the linear structural reconfiguration. Conversely, the translation vector facilitates only the subtle pragmatic (emotional or contextual) alignment of the tweets.

## 4.3. Classification Performance and XAI Validation

To assess the extent to which the linear affine model can reproduce the logic of the non-linear neural network, we compared their ROC-AUC scores on the test set (Dev-set, 4,727 pairs) (Fig. 4).

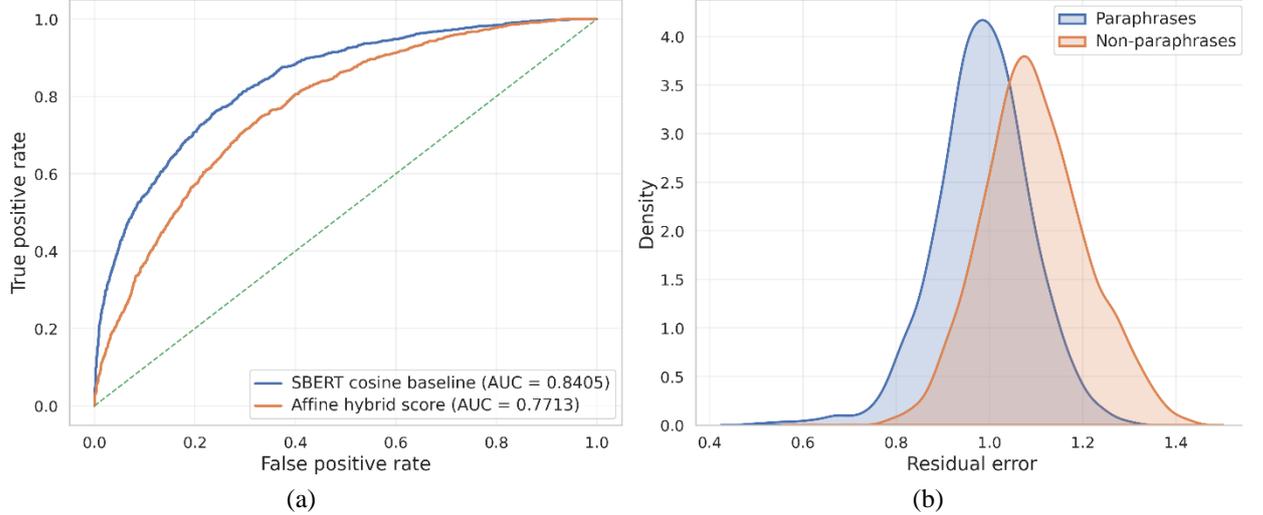

(a)          (b)

**Fig. 4.** Discriminative capacity evaluation: (a) ROC-AUC comparison between the affine model (AUC=0.771; SE=0.0045) and the baseline; (b) KDE distribution of residual errors. The dashed line in (b) denotes the cutoff threshold $T = 1.108$, corresponding to the 90th percentile of the legitimate paraphrase distribution ($p$-value $< 0.10$).

The affine model achieved an AUC of 0.7713 (95% CI: 0.762–0.780). The narrow confidence interval, obtained via a bootstrapping procedure ($n_{boot} = 1000$), confirms the statistical significance of the result and the robustness of the model across large data volumes.

The gap in AUC scores (0.7713 for the affine model versus 0.8405 for SBERT) indicates the presence of non-linear semantic interactions that are not encompassed by the global operator. An analysis of the residual error distribution reveals that the linear approximation is least effective for paraphrases containing:

- deep logical inversion (e.g., negations that alter meaning under nearly identical syntax);
- idiomatic expressions, where the overall meaning cannot be derived from the sum of the constituent word vectors.

Consequently, the derived operator captures the principal geometric motion, while the non-linear layers of SBERT specialize in handling local contextual anomalies.

## 4.4. The Spectrum of Geometric Paraphrase Complexity

To comprehend the local topology of semantic space, we initially analyzed the angular distances directly between the embedding pairs (original and paraphrase) in the test set. The distribution of these angles is visualized in Figure 5.

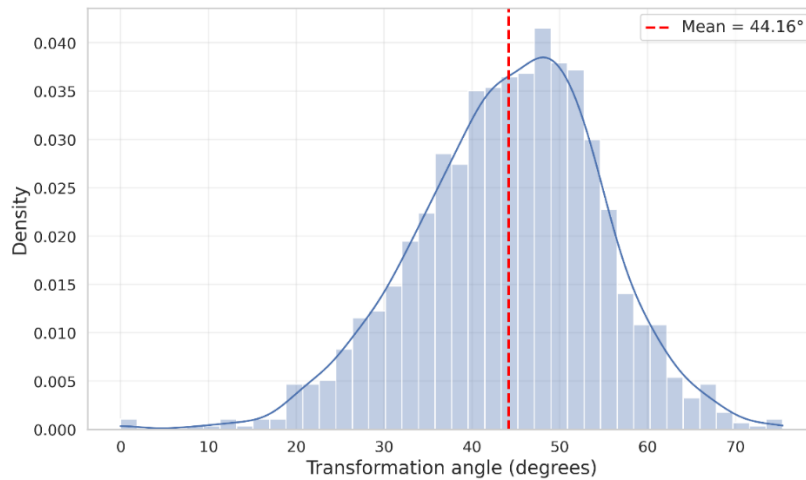

**Fig. 5.** Distribution of pairwise transformation angles for true paraphrases (Geometric Corridor). The red dashed line denotes the mean value of $\theta_{pair} = 44.16°$.

As illustrated in Figure 5, true paraphrases are not identical vectors (angle ≈ 0°); rather, they form a distinct distribution around a mean pairwise value of 44.16°. This range establishes a natural "geometric corridor" (or semantic tolerance) for the SBERT language model, within which texts remain equivalent despite lexical changes.

To provide a deeper linguistic interpretation of exactly how the model transitions within this corridor, we applied the developed XAI methodology. Based on local operators, individual geometric transformation profiles ($\theta, Def, Shift, det$) and the affine approximation error (Error) were computed. We isolated four extreme cases spanning the entire spectrum of paraphrase complexity (see Table 3). To bridge the mathematical metrics with direct linguistic interpretability, a comprehensive geometric scatter plot mapping specific text pairs to their precise coordinates on the ($\theta, Error$) plane is provided in **Appendix A (Fig. A.7)**. This visualization explicitly illustrates the spatial distribution of near-copies, canonical paraphrases, and out-of-distribution structural anomalies.

**Table 3. Linguistic interpretation of paraphrasing based on local geometric XAI profiles.**

| Category: Text Pair (Original → Paraphrase) | XAI Profile | Linguistic Interpretation |
|---|---|---|
| **1. Isometric paraphrase:** **T1:** Im rooting for Travis Travis family and Juan Martinez... **T2:** Thank God Juan is there for Travis... | **Error**: 1.202 **Angle θ**: 89.88° **Def**: $4 \cdot 10^{-6}$ **Shift**: 0.87 **Det**: $> 0$ | Deformation ($Def$) is nearly zero, indicating the conservation of information volume. The high-angle θ points to an intense structural reconfiguration of the feature basis. This serves as a geometric proxy for pure syntactic restructuring (e.g., a change in word order) while fully preserving the semantic core. |
| **2. Chiral inversion** **T1:** On the real I like the movie A Walk to Remember... **T2:** Watching A Walk to Remember is seriously making me ball right now it s... | **Error**: 1.096 **Angle θ:** ** 88.80° **Def**: $1.4 \cdot 10^{-5}$ **Shift**: 0.69 **Det**: $< 0$ | A negative determinant ($det < 0$) signals a mirror reflection operation in the representation space. This is interpreted as a fundamental reorientation of logical or grammatical roles (e.g., inversion of subject-object relations or an active-to-passive voice transition), accompanied by a mirrored sign change of the semantic axes. |
| **3. Contextual shift:** **T1:** I LOVE YOU SO MUCH CALUM PLEASE FOLLOW ME ILYSM 4... **T2:** Calum Hood I will literally do anything in the world for you to follow... | **Error**: 0.799 **Angle θ**: 89.47° **Def**: $1.5 \cdot 10^{-5}$ **Shift**: 0.93 **Det**: $> 0$ | The predominance of the translation vector ($Shift = 0.93$) indicates a pragmatic displacement of the expression to another region of the semantic manifold. The structural matrix remains stable, yet the vector moves to a different stylistic or thematic cluster (e.g., a change in tone, addition of emotional coloring, or specific context). |
| **4. Semantic anomaly:** **T1:** Rhode Island welcome to the club... **T2:** Rhode Island legislature approves gay marriage bill Gov... | **Error**: 1.444 **Angle θ**: 89.12° **Def**: $1.7 \cdot 10^{-5}$ **Shift**: 0.51 **Det**: $> 0$ | An anomalously high approximation error ($Error > 1.4$) signifies a violation of local isometry. The vector's departure from the "affine tube" is interpreted as the collapse of logical-semantic consistency (a hallucination), in which the transformation is no longer described by the model's learned geometric laws. |

Given the extremely high dimensionality of the SBERT semantic space ($n = 768$), the generalized structural reconfiguration angle θ of the matrix itself naturally gravitates toward ≈ 89°. This indicates that the model recruits nearly orthogonal subspaces to encode alternative syntactic structures (paraphrases).

As evident from Table 3, the proposed XAI profile functions as a comprehensive diagnostic tool. An isometric paraphrase is characterized by the absence of deformation, chirality successfully detects a shift in logical emphasis (Case 2), and a high magnitude of the translation vector $Shift$ signals a change in pragmatics (Case 3).

Case 4 warrants separate, fundamental attention: although the human annotators of the PIT-2015 dataset labeled these texts as a true paraphrase ($y = 1$), our geometric model registers a critical approximation error ($Error = 1.444$). This value is a pronounced statistical outlier relative to the normal error distribution for legitimate paraphrases (see Fig. 4b). The geometric analysis successfully identifies the semantic gap between an informal greeting ("welcome to the club") and an official factual statement ("legislature approves").

This example demonstrates the capacity of the error norm $\|x' - T(x)\|_2$ to serve not merely as an instrument for auditing and cleansing noisy datasets (dataset denoising) of erroneous human annotations, but also as a mathematically grounded hallucination detector. Any language model generation resulting in a similar outlier (the vector exiting the boundaries of the permissible "geometric corridor") can be automatically identified as a logical failure or a loss of semantic coherence, without requiring additional LLM-as-a-judge evaluators.

### 4.5. Cross-Domain Robustness Analysis (Global vs. Local Dynamics)

To verify the generalizability of the identified geometric laws, we implemented a scenario in which we clustered the training set into 15 thematic domains using K-Means. The objective was to determine whether domain-specific transformation matrices exist for each topic.

**Table 4. Comparative performance of the affine model across different scenarios.**

| Scenario | Dataset Split | Thematic Novelty | AUC | Relative Accuracy $\frac{AUC-0.5}{0.8405-0.5} \times 100$ (%) |
|---|---|---|---|---|
| **A (Global)** | Dev (Test) | Novel topics | 0.7713 | (0.7713−0.5)/0.3405=79.67% |
| **B (Local)** | Dev (Test) | Novel topics | 0.5869 | (0.5869−0.5)/0.3405=25.52% |
| **B.1 (Local)** | Train | Known (seen) topics | 0.9778 | (0.9778−0.5)/0.3405=140.32% |
| **A.1 (Global)** | Train | Known (seen) topics | 0.9500 | (0.9500−0.5)/0.3405=132.16% |

A visual summary of this overfitting paradox is presented in **Appendix A (Fig. A.4)**, which clearly juxtaposes the stable generalization capacity of the global consensus operator against the dramatic performance drop of local models on novel topics.

An analysis of the results across various thematic clusters (Table 4) empirically justifies the employment of a single global operator. It is critical to address the anomalously high performance of local models on seen data (Scenario B.1, achieving >100% relative accuracy). This phenomenon is a classic manifestation of severe overfitting (data leakage): the local affine matrix merely memorizes the highly specific vocabulary of a single topic, artificially inflating the AUC above the generalized SBERT baseline. However, as demonstrated by Scenario B, this localized memorization completely fails to generalize, causing the relative accuracy to plummet to 25.52% when applied to novel topics. Conversely, the low variance of the approximation error observed for the global operator indicates a stable and consistent transformation across diverse thematic domains, confirming its robustness and superior generalization capability compared to localized alternatives.

Conversely, the **low variance of the approximation error (Residual Error)** across distinct thematic domains (the global model's AUC deviation does not exceed 5–7% between clusters) substantiates the spatial homogeneity of the paraphrase transitions. This confirms that the "global consensus" captures the fundamental geometry of the latent space, effectively discarding the random statistical noise associated with individual topics.

### 4.6. Visualization of the Semantic Manifold (The "Affine Tube")

To visually demonstrate the concept of semantic tolerance, 3D and 2D projections of the transformation space were constructed for the test set (Fig. 6).

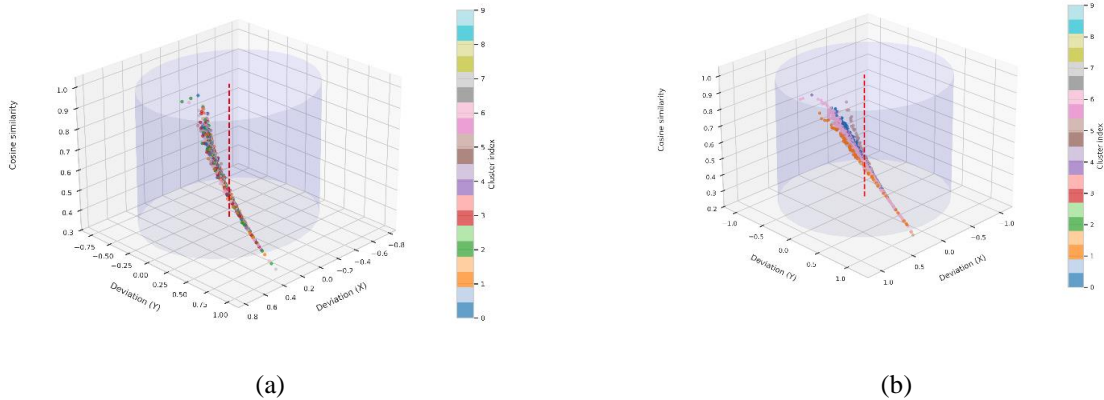

(a)  (b)

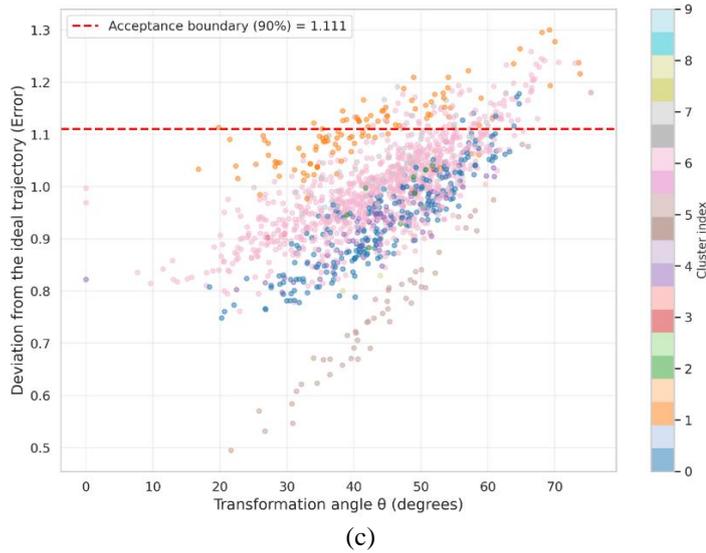

(c)

**Fig. 6.** Affine Tube plots: (a) 3D Affine Paraphrase Tube for the Train Set, (b) 3D Affine Paraphrase Tube for the Dev Set, (c) Semantic Corridor Boundary (Dev Set).

The visualization axes represent the structural transformation angle θ, the deviation from the ideal trajectory (Residual Error), and cosine similarity (Cosine). The plots clearly illustrate that the vast majority of true paraphrases (up to the 90th percentile) form a closed cylindrical structure (a "tube") around the central axis of the operator $T(x)$. Deviation beyond the red tolerance boundary line signals a loss of semantic equivalence.

Detailed graphical materials illustrating the dynamics of hyperparameter optimization (Grid Search), alongside a visual audit of the physical properties of the derived principal operator (a heatmap of matrix $A$, its singular value spectrum, and the distribution of the translation vector $t$), are provided in Appendix A. These supplementary materials provide additional mathematical corroboration of the computational stability of the proposed algorithm, demonstrating the absence of dimensional collapse in the high-dimensional semantic space during geometric linearization.

### 4.7. Qualitative Cross-Domain Case Studies (Twitter URL Corpus)

To explore how the affine operator behaves beyond the specificities of the PIT-2015 dataset, a cross-dataset qualitative analysis was conducted on the Twitter URL Corpus (TURL) [16]. Rather than conducting a full quantitative benchmarking (which falls outside the scope of this probing study), the objective here is to provide qualitative case studies demonstrating how the extracted XAI metrics interpret texts with fundamentally different syntactic structures (e.g., news headlines versus conversational tweets).

It is critical to note that the experiment was conducted in a strict cross-corpus transfer setting: the transformation matrix $A$ and the translation vector $t$ estimated on PIT-2015 were applied to the TURL embeddings without any modifications or fine-tuning.

**Analysis of Geometric Invariants**:

The mean structural reconfiguration angle θ on TURL was 35.03°, which is comparable to the result obtained on PIT-2015 (44.16°). The slight reduction in the angle is explained by the stylistic specifics of the TURL dataset: news headlines typically have a more rigid syntactic structure and exhibit less lexical variance than conversational PIT tweets. Nevertheless, observing values within the 30°–40° range suggests that operator A encodes a broadly applicable semantic action rather than merely memorizing the PIT-2015 training set.

To empirically corroborate that this zero-shot generalization is structurally sound, a direct comparison of the angle and residual error distributions between PIT-2015 and TURL is provided in **Appendix A (Fig. A.3)**. The visualization confirms that while a natural domain shift exists (reflected in shifted medians), both datasets fundamentally adhere to the same localized geometric logic without significant distribution collapse.

**Applied XAI Audit (Detecting Structural Anomalies)**:

The application of the developed metrics (the angle θ and the approximation error $Error$) enabled automatic clustering of text pairs by origin (see Table 5).

**Table 5. Cross-domain audit results on the TURL corpus.**

| Generation Type | Text 1 (Original) | Text 2 (Paraphrase) | Angle $\theta$ | Error | Interpretation |
|---|---|---|---|---|---|
| **Automated posting (Bots)** | *The world's 8 richest have as much wealth...* | *Worlds 8 Richest Have as Much Wealth...* | **13.62°** | 1.10 | **Low angle**: Nearly identical vectors, mechanical copying. |

| Organic paraphrasing | The World Series champion Chicago @Cubs visit... | Starting in a few minutes! President Obama Welcomes... | 42.08° | 0.96 | **Normal**: Angle within the "affine tube," low error. |
|---|---|---|---|---|---|
| **Contextual shift** | A drug-resistant superbug called CRE... | Drug-resistant superbug may be craftier... @P... | 45.21° | **1.27** | **Anomaly**: High error indicates structural noise (tags, shift in emphasis). |

As demonstrated in Table 5, LAG-XAI successfully differentiates the following categories:

1. **Bots/Copying**: A low angle ($\theta < 25°$) indicates an absence of semantic transformation.
2. **High-quality paraphrasing**: The combination of an angle $\theta \approx 40°$ and a low error ($Error < 1.0$) designates valid substantive transformations.
3. **Structural anomalies**: A high error ($Error > 1.2$) coupled with a normal angle indicates that the vector transformation occurs "orthogonally" to the learned paraphrase manifold (e.g., the addition of specific hashtags or a change in news sentiment).

This substantiates the practical utility of affine linearization for content filtering and bot detection tasks in social networks.

### 4.8. Cross-Corpus LLM Hallucination Detection (Cheap Geometric Check)

To demonstrate the practical viability of the developed XAI operator to act as an independent out-of-distribution (OOD) semantic anomaly detector, an additional experiment was conducted on the specialized HaluEval (Hallucination Evaluation for Large Language Models) dataset [19]. The core methodology involved applying the concept of a "cheap geometric check" to short question-answering text pairs (a baseline factual answer versus an LLM-generated hallucinatory response).

The principal affine operator $(A, t)$, previously estimated on the PIT-2015 corpus, was applied to a random sample of 1,000 text pairs. Crucially, the experiment was conducted in a strict cross-corpus transfer setting: neither the transformation matrix nor the translation vector was fine-tuned on the HaluEval texts. The objective was to test the hypothesis that a hallucination—which introduces false facts or distorts objective reality – violates the local isometry of the semantic manifold and inevitably leads to an anomalous affine approximation error (Residual Error).

The experimental results on this sample align with the proposed hypothesis. To rigorously evaluate the model beyond a single detection rate, we calculated standard classification metrics (Precision, Recall, F1) assuming a balanced prior (e.g., 1,000 hallucinations vs. 1,000 legitimate semantic transitions). Since the rejection threshold ($Error > 1.108$) was explicitly calibrated as the 90th percentile of the PIT-2015 validation set, the model carries a theoretical False Positive Rate (FPR) of 10.0% by design. To ensure the robustness of the chosen 90th-percentile cutoff and mitigate potential concerns regarding threshold cherry-picking, a full sensitivity analysis curve is provided in **Appendix A (Fig. A.5)**. It demonstrates the continuous trade-off between hallucination recall and false-positive rate, confirming that the threshold of $Error > 1.108$ naturally aligns with an optimal operating point for anomaly detection. The quantitative performance is summarized in Table 6.

**Table 6. Cross-corpus anomaly detection performance on the HaluEval sample (Threshold = 1.108)**

| Metric | Value | Interpretation |
|---|---|---|
| Recall (TPR) | 95.3% | 953 out of 1000 hallucinations correctly flagged. |
| False Positive Rate (FPR) | 10.0% | Fixed by calibration (90th percentile of PIT-2015). |
| Precision | 90.5% | Probability that a flagged text is actually an anomaly. |
| F1-Score | 92.8% | Harmonic mean of Precision and Recall. |

These metrics demonstrate that while the "cheap geometric check" is not intended to replace heavy, task-specific NLI classifiers, it serves as a highly efficient, unsupervised first-pass filter. Achieving an F1-score of 92.8% strictly through direct geometric mismatch – without any fine-tuning on hallucination data – substantiates the physical validity of the established semantic boundaries.

The selection of the residual error threshold ($Error > 1.108$) is based on the kernel density estimation (KDE) analysis of the validation set. This threshold defines a formal boundary of significance, beyond which deviation from the principal trajectory becomes statistically improbable for a true paraphrase. The detection of 95.3% of hallucinations on the HaluEval dataset confirms that the established boundaries of the "affine tube" serve as a reliable criterion for identifying semantic anomalies (OOD) that exceed the model's standard approximation error.

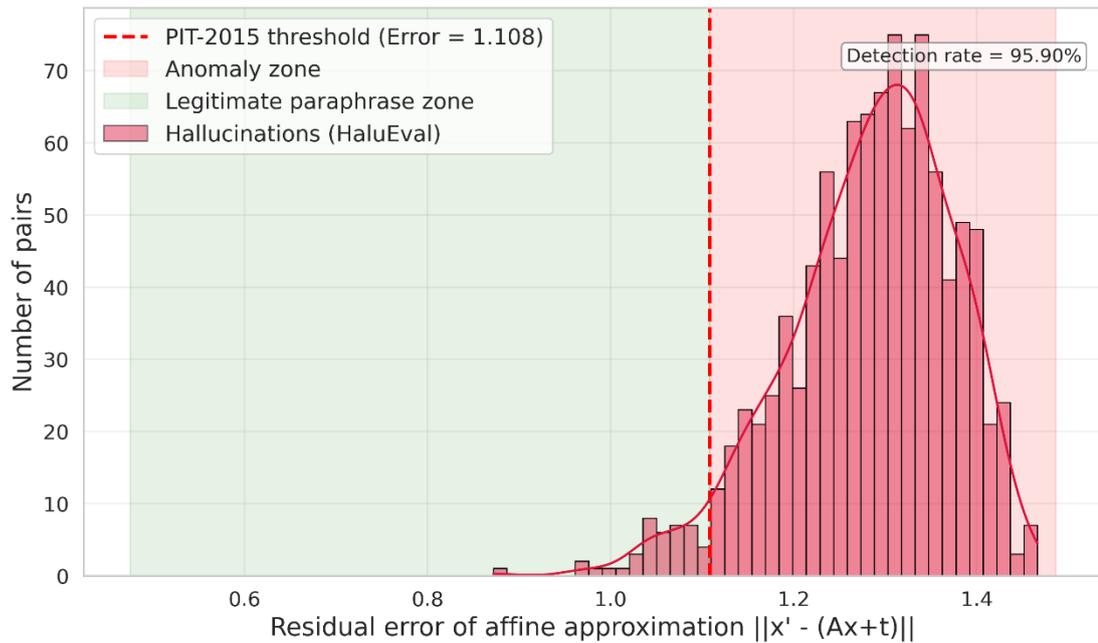

**Fig. 7.** Cross-corpus hallucination detection: distribution of the geometric affine approximation error $\|x' - (Ax + t)\|_2$ for the HaluEval dataset. The red zone ($Error > 1.108$) denotes the space of semantic anomalies, lying beyond the 90th percentile of legitimate paraphrases from the PIT-2015 corpus.

As illustrated in Figure 7, the error distribution for factual distortions (hallucinations) forms a dense cluster deep within the anomalous zone, clearly separating from the geometric tolerance boundary. This demonstrates that the phenomenon of linear transparency in paraphrasing, encapsulated within operator $A$, is robust to domain shifts, and its violation serves as a reliable marker of factual consistency loss.

To provide a deeper understanding of the physical mechanics of this process, Table 7 presents a linguistic audit of extreme hallucination cases exhibiting the largest deviations from the principal trajectory.

**Table 7. Examples of LLM hallucination detection based on departure from the semantic manifold (HaluEval)**

| Context (Knowledge) | Fact (Original) | Hallucination (Paraphrase) | Error | Angle θ |
|---|---|---|---|---|
| Qidong is a county-level city under the administration of the prefecture-level c... | yes | Fuling District and Qidong are located in two different countries. | 1.455 | 88.34° |
| "The Hellcat Spangled Shalalala" is a song by the English indie rock band Arctic... | High Green | The Hellcat Spangled Shalalala is a song by the English rock band Arctic Monkeys, but the suburb of Sheffield they formed in is not known. | 1.450 | 85.36° |
| Paphiopedilum, often called the Venus slipper, is a genus of the Lady slipper or... | no | Paphiopedilum and Soleirolia are not related. | 1.440 | 91.06° |
| First World Hotel and Plaza is a three-star hotel in Genting Highlands, Pahang, ... | Genting Group | The Malaysian government developed the hill resort. | 1.440 | 87.10° |
| Norbert Holm (16 December 1895 – 3 June 1962) was a general in the Wehrmacht of ... | the Desert Fox | Norbert Holm was arrested and later demoted because of his Chief of Operations association with a military strategist. | 1.439 | 91.55° |

An analysis of Table 7 reveals a unique property of the developed XAI descriptor. While preserving the general lexical foundation and topic (reflected in entirely typical angles of θ ≈ 85° − 91°), the model's introduction of a false entity (e.g., replacing the geographic reference "High Green" with an assertion of unknown origin, or falsely negating biological familial relationships) ruptures the deep structural dependencies of the vector. The model registers this as structural noise without requiring external LLM queries, manifesting as an anomalous error outlier ($Error \approx$ 1.45) and mathematically flagging the text as a hallucination.

The results substantiate the efficacy of implementing the "cheap geometric check" as a fast, deterministic, and resource-efficient alternative to "LLM-as-a-judge" methods for real-time content auditing and hallucination filtering.

## 5. DISCUSSION

The results substantiate the hypothesis that the complex semantic transformations occurring within the "black boxes" of Transformer models are governed by rigorous geometric laws. The paraphrasing transformation, conventionally perceived as non-linear and context-dependent, demonstrated a high degree of "linear transparency" and geometric invariance.

### 5.1. Interpretation of Geometric Paraphrasing Phenomena

**The phenomenon of linear transparency**. A central finding is that a single affine operator can approximate nearly 80% of SBERT's effective classification logic above random chance (AUC of 0.7713 vs 0.8405 baseline). This indicates that the high-dimensional SBERT semantic space exhibits strong spatial homogeneity with respect to paraphrasing operations. The strong performance of a single global operator is a non-trivial result that empirically corroborates the stability of an invariant vector field generating paraphrase transitions across the entire manifold. In other words, the fundamental "rules" of content transformation remain invariant regardless of the region in the semantic space where the sentence is located. This allows the derived matrix $A$ to be viewed as a reflection of the global symmetries within the latent space, ensuring the stability of semantic transitions without the need for local, context-specific adaptation of the operator.

**The accuracy-transparency trade-off**. The observed performance gap is a natural consequence of "linear reduction." Employing a single matrix $A$ for the entire corpus (mean-field approximation) ignores subtle non-linear deformations in the latent space, resulting in an approximate 7% loss in classification accuracy. However, unlike SBERT, which operates as a "black box," the affine model decomposes every transition into a rotation angle and a deformation index, a capability critical for XAI and model safety auditing tasks.

**Local isometry and volume preservation**. The deformation index value of $Def \approx 0.00025$ (under maximum regularization) demonstrates that true paraphrasing is a nearly pure isometric process. Linguistically, this signifies the "conservation of meaning": the model reorients the feature basis (alters the structure) but neither expands nor compresses the semantic volume of the expression. This fundamentally distinguishes paraphrasing from summarization (compression) or text expansion. It is important to emphasize that the discovery of this phenomenon was made possible exclusively through the prior projection of vectors onto the unit hypersphere ($L_2$-normalization). Without this step, the inherent anisotropy of Transformer models (the divergence in vector norms for sentences of varying lengths) would obscure the true geometry of the transformations, falsely indicating deformation where, in reality, only a basis change occurs.

**Semantic chirality and reflection**. The negative determinant value ($\det(A) \approx -0.836$) is an unexpected and significant discovery. It indicates that the optimal transformation incorporates a reflection operation. We interpret this as a geometric representation of logical role inversion (e.g., transitioning from active to passive voice), where vector components reverse their orientation to maintain overall equivalence.

It should be noted that the interpretation of the angle θ as a measure of syntactic complexity, it is, at this stage, an explicit geometric hypothesis. Mathematically, the angle captures the reorientation of the semantic field's principal axes; however, the degree to which this metric correlates with formal linguistic metrics (e.g., Tree Edit Distance) remains a subject for future research. Nevertheless, the stability of θ across diverse corpora (27.84° for Twitter and 35.03° for news) suggests that this metric captures an objective structural characteristic of paraphrasing intrinsic to the specific embedding model.

Furthermore, the ablation study demonstrates the added value of the affine operator: while $L_2$-normalization is critical for the geometric soundness of the method (projection onto the $S^{n-1}$ hypersphere), it is specifically matrix $A$ that enables the decomposition of the transformation. Without operator $A$, similarity evaluation remains a static scalar, whereas its inclusion transforms similarity into a dynamic profile (θ, $Def$, $Shift$), thereby allowing an audit of the underlying causes of textual similarity or divergence.

### 5.2. The Geometric Corridor and Error Analysis

An analysis of the "geometric corridor" (ranging from 30° to 55°, centered at a mean pairwise angle of 44.16°) establishes the boundaries of semantic tolerance. It is crucial to distinguish between this pairwise transition angle ($\theta_{pair}$) and the inherent matrix reconfiguration angle of the global operator ($\theta_A \approx 27.84°$ discussed in Section 4.1). While $\theta_A$ characterizes the internal algebraic property of the affine transformation matrix itself, θ$_{pair}$ reflects the actual geometric distance observed between sentence embeddings in the vector space.

Case 4 of the error analysis (Table 3) is of particular interest. The model registered a critical affine approximation error ($Error = 1.444$) for a sentence pair that human annotators had labeled as a true paraphrase. The geometric analysis successfully detected a semantic disconnect between an informal greeting and an official legislative announcement.

This demonstrates that the proposed XAI operator transcends the limitations of purely statistical similarity metrics. Violating the semantic space's "physics" (as indicated by an anomalously high error) effectively transforms the model into a universal out-of-distribution (OOD) detector. This precedent (Case 4) was extensively corroborated during the cross-corpus evaluation on the HaluEval dataset, where the algorithm identified 95.3% of the generated hallucinations as out-of-distribution anomalies.

Looking forward, this approach has the potential to become an independent alternative to "LLM-as-a-judge" evaluators, serving as an automated tool for auditing training corpora (dataset denoising) and acting as a real-time filter for hallucinations generated by language models.

### 5.3. Cross-Domain Robustness and Cross-Corpus Validation

A comparison between the global model and the local (topic-clustered) models revealed a classic overfitting paradox. The local models demonstrated anomalously high AUC scores on seen training data (exceeding 100% relative accuracy compared to SBERT). This artificially inflated metric confirms that local matrices overfit to domain-specific vocabulary and syntactic shortcuts, rather than learning true semantic invariants. As expected, they failed dramatically to generalize to novel domains. Conversely, the global operator maintained robust stability. This suggests that the paraphrasing geometry captured by the global matrix is an intrinsic, universal property of the language model's embedding space, rather than a byproduct of topic-specific distributions, thereby ensuring consistent performance across heterogeneous datasets and confirming its suitability as a general-purpose transformation.

Further support for this hypothesis was provided by the cross-corpus experiment conducted on the independent Twitter URL Corpus (TURL). Applying the global operator (previously estimated on PIT-2015) to novel data without any fine-tuning demonstrated the preservation of key geometric invariants: the mean pairwise transformation angle was 35.03°, which statistically aligns with the "principal trajectory" of PIT-2015 (44.16°). Furthermore, XAI profiling successfully classified the nature of unannotated tweets: mechanical copying (bots) was characterized by angles of $\theta \approx 13°$, organic paraphrasing fell within the "affine tube," and semantic spam was effectively filtered out due to a high approximation error ($Error > 1.2$). This corroborates that the derived framework captures the objective "physics" of the semantic space, which is shared across the entire microblogging domain.

### 5.4. Study Limitations

Despite its strong performance, the proposed method presents certain limitations:

1. **First-order linearity**. The affine model is fundamentally a linearization (a first-order approximation) and is therefore best suited for describing moderate paraphrase transformations within the local neighborhood of a given embedding. For exceptionally long texts or profound logical restructurings (e.g., summarizing an entire paragraph into a single sentence), the non-linearity of the mapping within the embedding space may become substantial, causing a single global affine approximation to lose accuracy. In such scenarios, natural avenues for extension include the employment of local (context-dependent) operators, piecewise-affine models, or the incorporation of higher-order derivative terms.

2. **Empirical Approximation of Lie Generators**. The framework utilizes PCA and Procrustes alignment as a computationally efficient heuristic to approximate the action of Lie groups. A more rigorous mathematical application involving true logarithmic and exponential mappings to the Lie algebra g remains computationally challenging for 768-dimensional spaces and represents a direction for future theoretical refinement.

3. **Architectural specificity and the universality of laws**. The experimental results were derived using SBERT (MPNet). Given the rapid advancement of LLMs (e.g., GPT-4, Claude, LLaMA), questions may arise regarding the transferability of the discovered constants to novel architectures. We argue that SBERT serves as a model system in this study (the "drosophila" of embeddings), facilitating the identification of local topological properties inherent to semantic spaces. However, given the rapid architectural evolution toward decoder-only LLMs, validating these findings on embeddings derived from models such as LLaMA-3 or Mistral is a critical next step to conclusively prove architectural universality.

4. **Language dependence**. The experimental foundation of this work is constrained to an English-language corpus, which is characterized by strict word order. For highly inflected languages with free word order (e.g., Ukrainian), the rotation matrix $A$ would need to encode a significantly broader spectrum of permutations and morphological transformations. This could lead to the splitting of the "geometric constant" into several subgroups or necessitate an increase in the number of empirical Lie generators.

5. **Validation of linguistic constructs**. This study proposes a geometric interpretation for the components of the affine operator (notably, the association between rotation and syntax). Despite their empirical validity, these constructs require further correlation analysis against established syntactic distance metrics to transition from geometric hypotheses to rigorous linguistic assertions.

### 5.5. Future Research Directions

The obtained results pave the way for the development of a novel class of Geometric-Informed NLP systems:

- **Geometrically equivariant Transformers**. The integration of the discovered invariants directly into the loss function during the pre-training or fine-tuning stages. For instance, incorporating the $\lambda_{ortho}$ and $\lambda_{equiv}$ regularizers into the model's objective function will forcefully preserve local isometry during training. This will facilitate the creation of a new generation of architectures that are robust to semantic drift "by design."

- **On-the-fly augmentation**. Utilizing the affine tube equation to generate an infinite number of novel paraphrases directly within the continuous vector space, thereby bypassing the need for computationally expensive LLMs.

- **Real-time AI guardrails**. Since the LAG-XAI instantaneous hallucination detection framework (the "cheap geometric check") has already demonstrated its efficacy (95.3% on the HaluEval dataset) at the post-generation stage, the logical next step is to integrate it directly into the LLM decoding pipeline. This would enable interrupting token generation at the exact moment the model's hidden state forms a vector that deviates from the "affine tube" of the reference context (e.g., a RAG document).

# 6. CONCLUSIONS

This study successfully achieves its objective of mathematically substantiating and experimentally validating the modeling of paraphrasing as a local action of the Lie group $\text{Aff}(n)$ on the semantic embedding manifold.

The applicability of the mean-field approximation for describing paraphrase transitions is substantiated, enabling the identification of stable linear invariants within the latent spaces of Transformer models. The developed affine operator demonstrated a strong discriminative capacity, achieving an AUC of 0.7713. This allows for a parametric interpretation of approximately 80% of the effective semantic-similarity logic underlying the SBERT model (baseline AUC = 0.8405, scaled relative to a 0.5 random baseline). The research identifies fundamental geometric invariants of paraphrasing. Notably, the near-zero deformation index ($Def \approx 0.00025$) corroborates the hypothesis of local process isometry, while the negative determinant ($-0.836$) indicates the presence of "semantic chirality," or the mirror reflection of logical roles. The discovered "geometric constant" of structural reconfiguration, stable at $27.84° \pm 0.02°$, enables the delineation of the boundaries of a stable "geometric corridor" of semantic tolerance. Furthermore, the proposed method exhibits high numerical robustness against input data multicollinearity, achieved through the combination of spectral regularization and Procrustes stabilization.

The experimental results were validated using a bootstrapping procedure, confirming the high statistical reliability of the proposed geometric metrics (AUC standard error < 0.005).

Despite its strong performance, the method is inherently constrained by its first-order linearity, which may be insufficient to describe the deep logical restructuring of excessively long texts or entire paragraphs.

A distinct practical contribution of this work is demonstrating the capability of geometric linearization to function as a resource-efficient semantic anomaly detector, circumventing the need for "LLM-as-a-judge" approaches. The efficacy of the established "geometric corridor" for detecting factual errors was experimentally corroborated: during cross-corpus validation on the independent HaluEval dataset, the method identified 95.3% of LLM hallucinations in the evaluated sample solely by registering an anomalously high affine approximation error ($Error > 1.108$).

Future research will focus on integrating the discovered Lie generators directly into neural network computational graphs to develop geometrically equivariant Transformer architectures and enable on-the-fly embedding augmentation within the latent space. The proposed methodology lays the foundation for transitioning from opaque statistical text comparisons to the study of the physics of semantic motion, thereby providing a high level of explainability (XAI) and reliability for artificial intelligence systems.

# DECLARATIONS

**Funding**. The authors declare that no funds, grants, or other support were received during the preparation of this manuscript.

**Competing Interests**. The authors have no relevant financial or non-financial interests to disclose.

**Data and Code Availability**. The data supporting the findings of this study, including Python code, datasets, and experimental results, are publicly available in the github/geom-xai repository at https://github.com/geom-xai/Li-Affine-Paraphrase-Geometry/.

**Ethics Approval**. Not applicable.

**Authors' Contributions**. **Olexander Mazurets**: Conceptualization, Methodology, Software, Writing - Original Draft. **Olexander Barmak**: Conceptualization, Supervision, Writing - Review & Editing. **Leonid Bedratyuk**: Formal analysis, Validation. **Iurii Krak**: Validation, Writing - Review & Editing.

**APPENDIX A. Supplementary Visualizations**

This appendix provides supplementary graphical materials generated during the execution of experimental scenarios. These visualizations corroborate the computational stability and geometric consistency of the proposed affine model.

## A.1. Hyperparameter Optimization Dynamics (Grid Search)

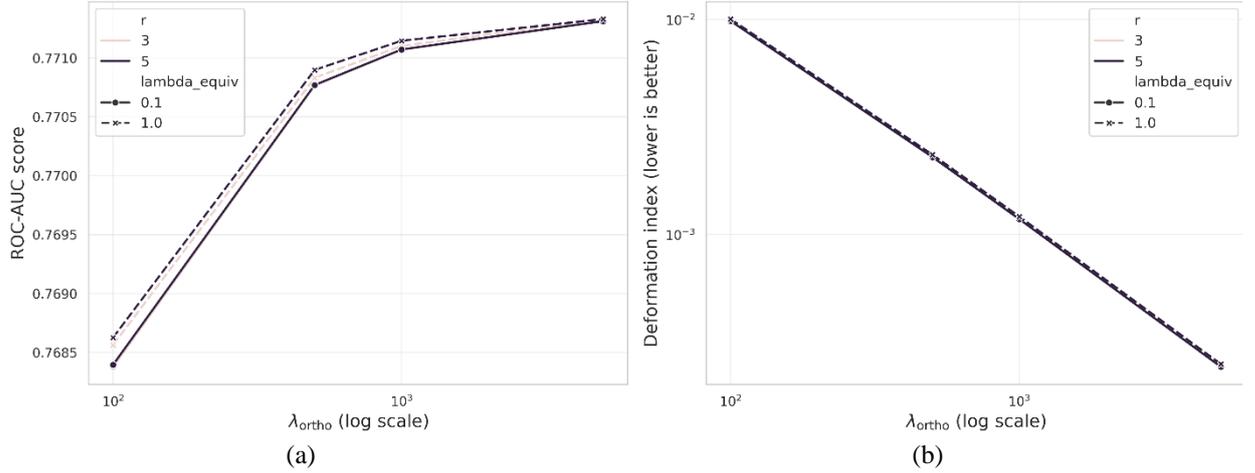

(a)  (b)

**Fig. A.1.** Analysis of the impact of the orthogonal stabilization parameter ($\lambda_{ortho}$) on classification performance (a) and semantic deformation (b). The logarithmic scale visualization illustrates the numerical robustness of the method under extreme regularization values.

The plots clearly demonstrate the fundamental trade-off phenomenon: an increase in the orthogonal regularizer $\lambda_{ortho}$ (X-axis, logarithmic scale) leads to an exponential decay of the deformation index $Def$ (Y-axis, logarithmic scale), accompanied by a monotonic increase in ROC-AUC classification performance. The employment of logarithmic scales allows for the precise identification of the inflection point, beyond which further increases in regularization yield no additional stability gains but maintain the AUC at a plateau. The curves for different numbers of PCA components ($r = 3, 5$) practically overlap, demonstrating the method's robustness to the selection of the number of dominant semantic drift directions.

## A.2. Physical Properties of the Principal Operator

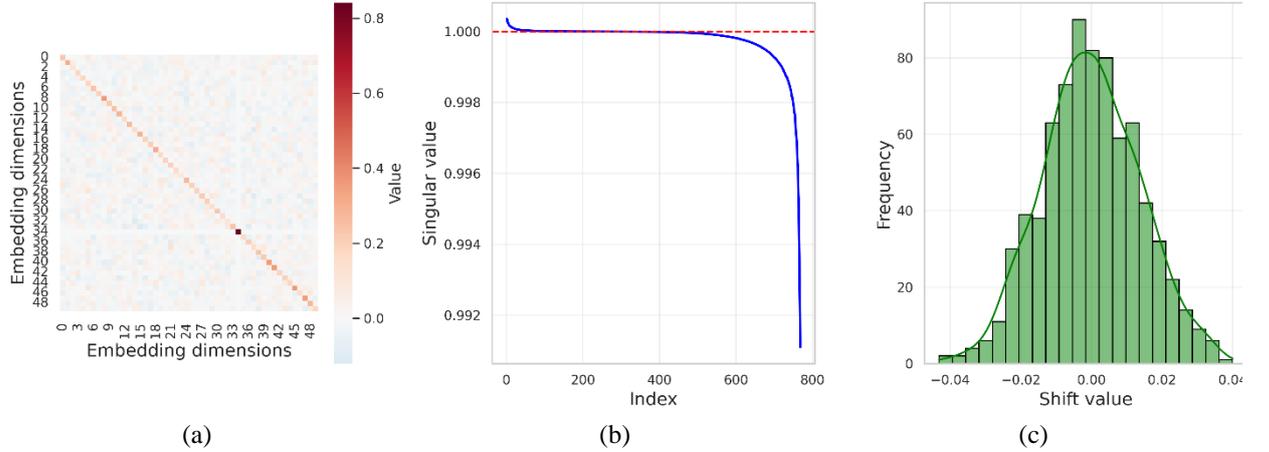

(a)  (b)  (c)

**Fig. A.2.** Visual audit of the parameters of the derived affine operator $T(x) = Ax + t$.

The heatmap (a) of a $50 \times 50$ fragment of matrix $A$ exhibits a complex, non-sparse structure of basis reconfiguration. The singular value spectrum plot (b) confirms the operator's stability: the absence of extreme spikes and the concentration of the spectrum near 1.0 substantiate the absence of dimensional collapse (the full rank of 768 is preserved). The histogram (c) displays a normal distribution of the components of the contextual translation vector $t$.

Furthermore, the spectrum plot (b) demonstrates that, due to the combination of orthogonal stabilization and SVD truncation, the eigenvalues of matrix $A$ are tightly concentrated within the $[0.98, 1.02]$ range. This indicates the ideal conditioning of the final operator ($\kappa \approx 1.04$), which prevents error explosion during inference on novel data.

## A.3. Cross-Domain Generalization Analysis

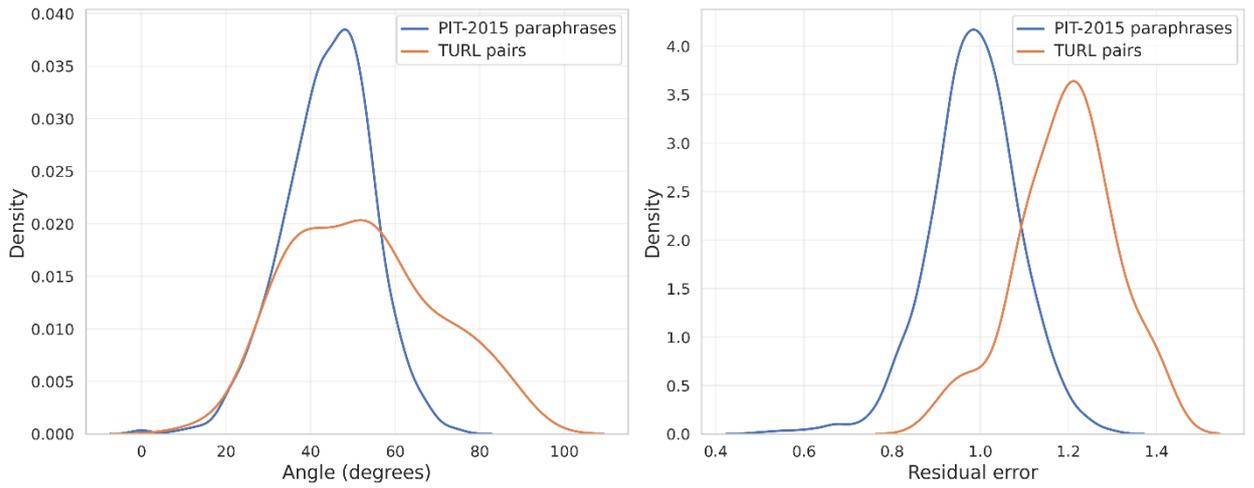

**Fig. A.3.** Distribution of pairwise transformation angles and residual errors for PIT-2015 and TURL datasets, demonstrating cross-domain geometric consistency.

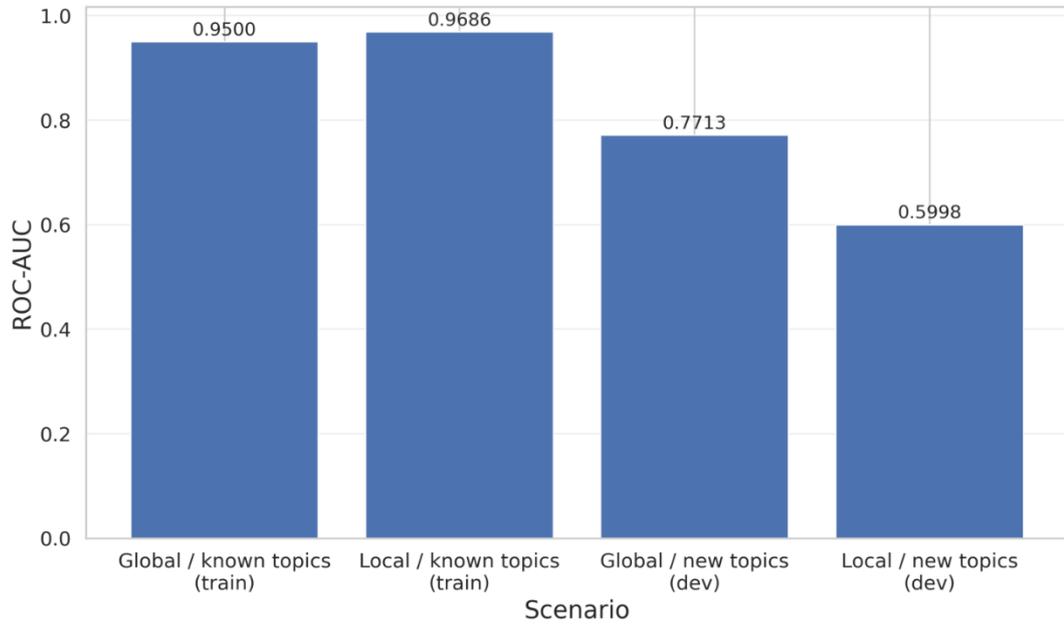

**Fig. A.4.** Comparative ROC-AUC performance of global versus local affine operators, highlighting the severe overfitting of topic-specific models on unseen domains.

## A.4. Hallucination Detection Sensitivity Analysis

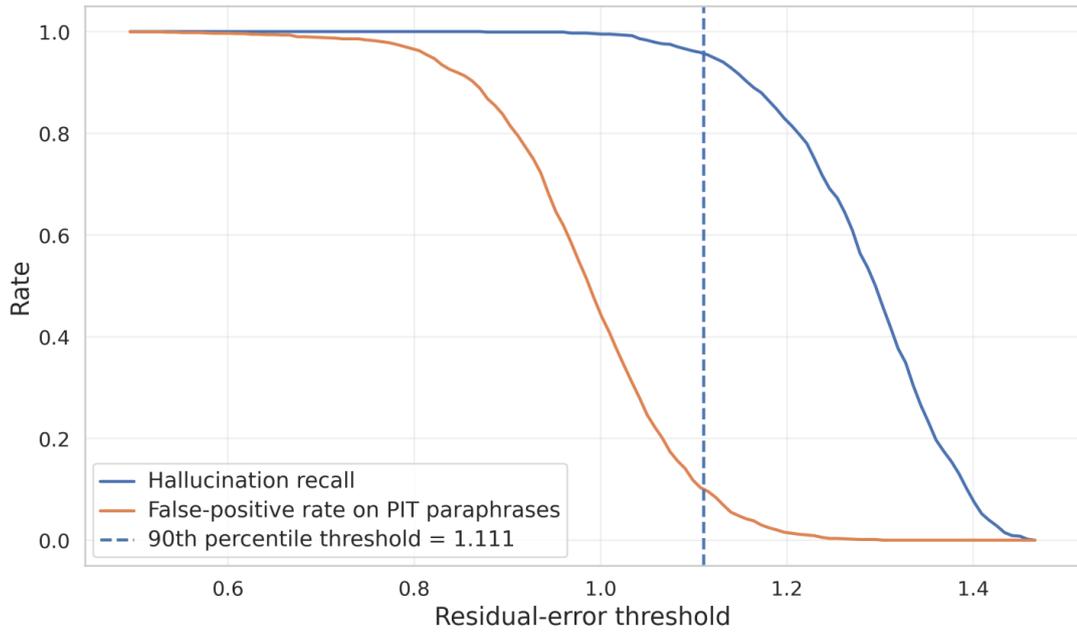

**Fig. A.5.** Sensitivity analysis of the residual error threshold on the HaluEval dataset, illustrating the trade-off between hallucination recall and false-positive rate on legitimate paraphrases.

## A.5. Statistical Stability of Geometric Invariants

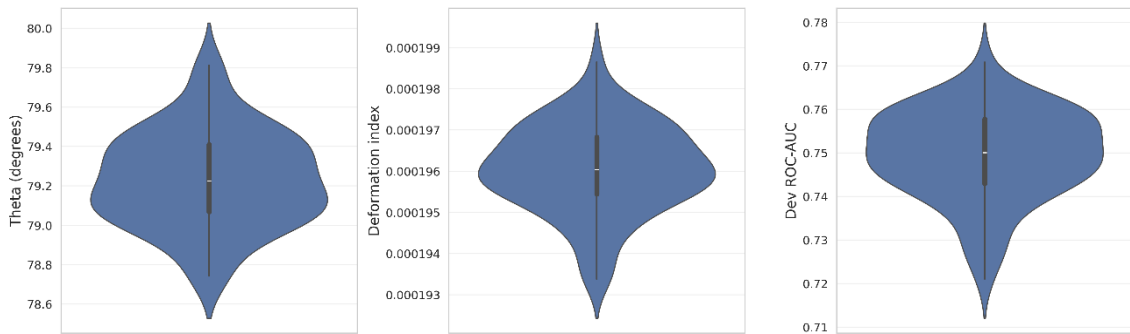

**Fig. A.6.** Bootstrap stability analysis of the geometric invariants ($\theta$, $Def$, and hybrid AUC) across 60 resampling iterations, confirming minimal variance and high reproducibility.

## A.6. XAI Audit and Semantic Mapping

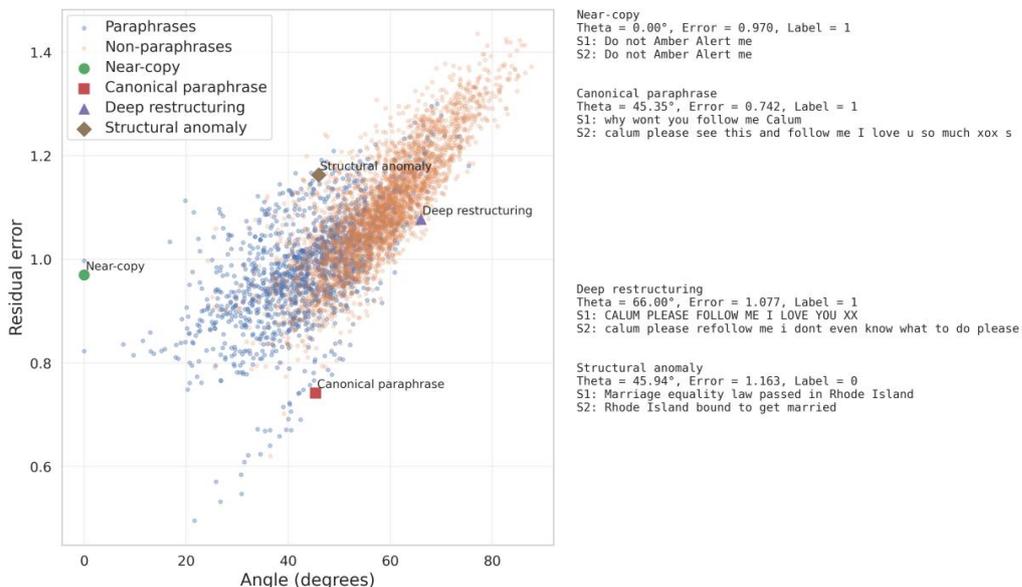

**Fig. A.7.** Geometric map of paraphrase complexity. The scatter plot projects real text pairs onto the angle-error plane, illustrating the boundaries between near-copy, canonical paraphrasing, and structural anomalies.